\pdfoutput=1
\documentclass[11pt]{article}
\usepackage[margin=1in]{geometry}
\usepackage[T1]{fontenc}
\usepackage[utf8]{inputenc}
\usepackage{times}
\usepackage{microtype}
\usepackage{amsmath,amssymb,amsthm,bm}
\usepackage{mathtools}
\usepackage{graphicx}
\usepackage{booktabs}
\usepackage{adjustbox}
\usepackage{algorithm}
\usepackage{algorithmic}
\usepackage[round]{natbib}
\usepackage{caption}
\usepackage{url}
\usepackage[dvipsnames]{xcolor}
\usepackage[colorlinks=true,linkcolor=MidnightBlue,citecolor=MidnightBlue,urlcolor=MidnightBlue]{hyperref}

\theoremstyle{plain}
\newtheorem{theorem}{Theorem}
\newtheorem{proposition}{Proposition}
\newtheorem{lemma}{Lemma}
\newtheorem{corollary}{Corollary}
\newtheorem{conjecture}{Conjecture}
\theoremstyle{definition}

\newtheorem{assumption}{Assumption}
\newtheorem{remark}{Remark}
\newcommand{\R}{\mathbb{R}}
\newcommand{\E}{\mathbb{E}}
\newcommand{\PP}{\mathbb{P}}
\newcommand{\Var}{\mathrm{Var}}

\newcommand{\nuhat}{\widehat{\nu}}
\newcommand{\fbar}{\bar f}
\newcommand{\indc}[1]{\mathbf{1}\!\left\{#1\right\}}
\newcommand{\xv}{\bm{x}}
\newcommand{\Xb}{\bm{X}}
\newcommand{\yv}{\bm{y}}

\newcommand{\abstracttext}{%
Automatic relevance determination (ARD), the default tool for variable selection in Gaussian-process (GP) regression, ranks inputs by inverse lengthscales---which measure how fast a function varies, not how much an input contributes to prediction---and offers no calibrated rule for deciding which inputs to keep. The prediction-centred alternative, the derivative sensitivity $\nu_j=\E\{(\partial f/\partial x_j)^2\}$, is available in closed form from a fitted GP, but turning it into a selection rule is harder than it looks: at a null input the estimator is a degenerate quadratic form, so Wald and Bernstein--von Mises cutoffs are anti-conservative, and the natural residual bootstrap is mis-scaled. We show that a studentized multiplier bootstrap of the GP derivative process repairs both, prove its validity through an invariance principle for quadratic forms, and obtain asymptotic family-wise and false-discovery-rate control across inputs. Over $100$ replications the rule controls FDR wherever inputs are truly null, while uncalibrated derivative rankings breach the target by up to $2\times$ and a Bernstein--von Mises cutoff by $2.2\times$; at matched FDR it loses no power; it holds under a Mat\'ern kernel and input correlation up to $0.99$; on real data with planted and authentic null inputs it admits $5$--$12\times$ fewer spurious inputs; it costs $5$--$18\%$ of the GP fit; and a block-averaged variant retains validity at cost linear in $n$.%
}

\title{Calibrated Derivative-Process Sensitivity for\\ Gaussian-Process Variable Selection}
\author{Jia Cai\\[2pt] Department of Statistics, George Mason University\\ Fairfax, Virginia, USA\\ \texttt{jcai8@gmu.edu}}
\date{}

\begin{document}
\maketitle

\begin{abstract}
\abstracttext
\end{abstract}

\section{Introduction}\label{sec:intro}

Gaussian processes (GPs) are a workhorse for regression and surrogate modeling, and the standard route to variable selection is \emph{automatic relevance determination} (ARD): fit an anisotropic kernel with a separate lengthscale $\ell_j$ per input coordinate, and read off the inverse lengthscales $1/\ell_j^2$ as relevance scores \citep{neal1996bayesian,rasmussen2006gpml}. ARD is attractive because the lengthscales come for free with the fit. But it has two long-recognized defects.

\textbf{1. The lengthscale measures the wrong thing.} The inverse lengthscale captures the \emph{rate of variation} of $f$ along coordinate $j$, not the \emph{magnitude of that coordinate's contribution to the prediction}. A coordinate that enters $f$ linearly with a large coefficient varies slowly (small $1/\ell_j^2$), while a coordinate that enters with a tiny but high-frequency oscillation varies rapidly (large $1/\ell_j^2$). ARD therefore ranks the second above the first even when the first dominates the predictive surface---the \emph{linear-versus-nonlinear misranking} documented by \citet{paananen2019variable} and \citet{piironen2016projection}.

\textbf{2. There is no calibrated decision rule.} Even granting a relevance ranking, ARD provides no principled cutoff. Practitioners threshold at a visual ``elbow'' or append synthetic inert reference inputs and keep coordinates whose lengthscale beats the references---an ad hoc device with no error guarantee.

A natural fix for the first defect is to score coordinates by a \emph{prediction-centered} quantity. The derivative-based global sensitivity measure
\begin{equation}\label{eq:nu}
  \nu_j \;=\; \E_{\xv\sim P_X}\!\left[\Big(\tfrac{\partial f}{\partial x_j}(\xv)\Big)^2\right]
\end{equation}
measures the average squared sensitivity of $f$ to $x_j$ and tracks amplitude $\times$ variation rather than variation alone. Because differentiation is linear, the derivative of a GP is itself a GP \citep{rasmussen2006gpml}, so $\nu_j$---the $j$-th diagonal of the expected-gradient-outer-product (``active subspace'') matrix \citep{constantine2015active}---is available in closed form from a fitted GP \citep{delozzo2016estimation,wycoff2021sensitivity}. This is essentially the score behind the sensitivity ranking of \citet{paananen2019variable}. Yet in all of this work $\nu_j$ is used only as a \emph{heuristic ranking}: there is no consistency theory, no test of whether a coordinate matters, and---critically---no calibrated cutoff. The second defect remains open.

\textbf{Contributions.} The functional $\nu_j$, its closed form under a GP, and even its asymptotic normality are known \citep{delozzo2016estimation,wycoff2021sensitivity,deng2022unified}, and per-coordinate tests of ``is a derivative zero'' exist for splines and kernel ridge regression \citep{racine1997consistent,liu2023estimation}. What no prior work supplies is a valid, \emph{simultaneous} selection rule, and we show that the obvious routes to one fail for a structural reason:
\begin{itemize}\setlength{\itemsep}{1pt}
\item \textbf{The null is degenerate (\S\ref{sec:null}).} At a null coordinate the first-order term of $\nuhat_j$ vanishes and the estimator is a positive weighted-$\chi^2$ form, so every Wald or Bernstein--von Mises interval---including the BvM of \citet{deng2022unified}---is mis-calibrated at exactly the coordinates FDR is about. We verify this: a Wald/BvM cutoff breaches the FDR target by $1.5$--$2.2\times$ (Table~\ref{tab:necessity}).
\item \textbf{The natural bootstrap is mis-scaled, and studentization repairs it (\S\ref{sec:errctrl}).} The raw residual multiplier bootstrap underestimates the null law by the residual-shrinkage factor $\gamma_n=\mathrm{tr}\{(I-\mathsf S)(I-\mathsf S)^\top\}/n$ (Kolmogorov distance $\approx0.3$); the studentized version matches it ($\approx0.02$). We prove this via an invariance principle for quadratic forms with an explicit spectral condition (Lemma~\ref{lem:boot}).
\item \textbf{Simultaneous error control (\S\ref{sec:errctrl}).} A joint limit for the vector of studentized statistics gives FWER control via the max statistic and, through super-uniform null p-values, FDR control via a dependence-robust step-up (Theorem~\ref{thm:boot})---the first such guarantee for derivative-based GP relevance.
\item \textbf{Two consequences of the prediction-centred functional.} It provably reverses ARD's misranking (\S\ref{sec:method}) and is stably estimable across the signal-variance/lengthscale ridge where individual lengthscales are not (\S\ref{sec:identif}).
\item \textbf{Evidence (\S\ref{sec:experiments}).} Over $100$ replications DPS controls FDR wherever coordinates are truly null while the uncalibrated derivative rankings breach by up to $2\times$; at matched FDR it loses no power; on real data with planted and authentic nulls it admits $5$--$12\times$ fewer spurious inputs; and a real-data stress test exposes an overfit regime in which the in-sample bootstrap fails and leave-one-out residuals repair it.
\end{itemize}

\textbf{Relation to conformal/e-value GP selection.} A complementary line obtains \emph{finite-sample} FDR control for the \emph{lengthscale} ranking via exchangeability of synthetic references and (e-)Benjamini--Hochberg. Our target is different (a prediction-centered functional that fixes the misranking) and our toolkit is disjoint (debiasing + multiplier bootstrap, not conformal p-values). The price of the disjoint toolkit is that our guarantee is asymptotic rather than finite-sample; we state this trade-off honestly.

\section{Related Work}\label{sec:related}

\textbf{Relevance in GPs.} ARD lengthscales as importance scores go back to \citet{neal1996bayesian} and are standard practice \citep{rasmussen2006gpml}. Recognizing that lengthscales conflate variation with contribution, \citet{paananen2019variable} proposed ranking inputs by the sensitivity of the posterior predictive distribution, and \citet{piironen2016projection} by projection-predictive model selection; both are heuristic rankings without a calibrated cutoff or error control. Sparse axis-aligned priors \citep{eriksson2021saasbo} shrink lengthscales for high-dimensional Bayesian optimization but again provide no testing. Our score is the derivative-based sensitivity behind these ideas, made into a calibrated test.

\textbf{Derivative processes and active subspaces.} The derivative of a GP is a GP \citep{rasmussen2006gpml}, and the expected-gradient-outer-product matrix---the active subspace \citep{constantine2015active}---has a closed form under a GP, used for sensitivity estimation by \citet{delozzo2016estimation} and \citet{wycoff2021sensitivity}, for active learning by \citet{belakaria2024active} and \citet{lambert2026gradient}, and, closest to us, by \citet{deng2022unified}, who derive the posterior of the same functional and a Bernstein--von Mises theorem for it but select by ranking rather than by a test. Plug-in GP derivatives attain optimal contraction rates \citep{liu2026optimal}, which supports our Assumption~\ref{ass:reg}(A3). Table~\ref{tab:contrast} places these against ours.

\begin{table*}[t]
\centering\footnotesize
\caption{Closest prior work. ``Null'': analysis of the degenerate null; ``Boot.'': a valid bootstrap for it; ``Sim.'': simultaneous FWER/FDR control over coordinates.}
\label{tab:contrast}
\begin{tabular}{l c c c c c}
\toprule
 & GP & $\nu_j$ closed form & test $\nu_j{=}0$ & Null/Boot. & Sim.\\
\midrule
\citet{wycoff2021sensitivity} & \checkmark & \checkmark & -- & -- & --\\
\citet{delozzo2016estimation} & \checkmark & \checkmark & per-input & -- & --\\
\citet{deng2022unified} & \checkmark & \checkmark & BvM interval & -- & --\\
\citet{liu2023estimation} & KRR & -- & per-derivative & boot. & --\\
\citet{paananen2019variable} & \checkmark & ranking & -- & -- & --\\
\citet{dai2022kernel} & additive & -- & knockoffs & -- & FDR\\
This paper & \checkmark & \checkmark & studentized boot. & \checkmark & FWER+FDR\\
\bottomrule
\end{tabular}
\end{table*}

\textbf{Quadratic functionals and significance testing.} $\nu_j$ is a quadratic functional of the derivative of $f$, the class of \citet{bickel1988estimating} and \citet{robins2017higher}, whose ``elbow'' phenomenon we inherit; average-derivative functionals are root-$n$ estimable \citep{hardle1989investigating,newey1993efficiency}. Testing whether a covariate matters in nonparametric regression has a long history \citep{racine1997consistent,lavergne2015significance,liu2023estimation}, and multiplier bootstraps for kernel-ridge \emph{function} bands exist \citep{singh2023kernel}; none addresses a degenerate derivative functional under a GP nor multiplicity across inputs.

\textbf{Error-controlled selection.} Benjamini--Hochberg \citep{benjamini1995controlling} and its dependence-robust form \citep{benjamini2001control} are our step-up engines; simultaneous calibration uses the high-dimensional bootstrap for degenerate $U$-statistics \citep{chen2018gaussian,chen2019randomized}. Model-X knockoffs \citep{candes2018panning,barber2015controlling}, kernel knockoffs for additive models \citep{dai2022kernel}, HSIC screening \citep{daveiga2015global}, RATE \citep{crawford2019rate,crawford2018bayesian}, and the Sobol'-index test of \citet{klein2022significance} are alternatives; Bayesian selection consistency for GPs is studied by \citet{jiang2021variable}. A complementary route obtains finite-sample FDR for the lengthscale ranking via exchangeability and (e-)BH \citep{wang2022false}; we use a disjoint toolkit and target a different, prediction-centered functional.

\section{Method}\label{sec:method}

\textbf{Setup.} We observe $(\xv_i,y_i)_{i=1}^n$ with $\xv_i\in\R^D$ drawn i.i.d.\ from $P_X$ and $y_i=f(\xv_i)+\varepsilon_i$, $\varepsilon_i\sim\mathcal N(0,\sigma^2)$. We place a GP prior $f\sim\mathcal{GP}(0,k)$ with the squared-exponential ARD kernel $k(\xv,\xv')=\eta\exp(-\tfrac12\sum_d (x_d-x'_d)^2/\ell_d^2)$ and fit $(\bm\ell,\eta,\sigma^2)$ by maximizing the marginal likelihood. Write $K$ for the kernel matrix, $K_n=K+\sigma^2 I$, and $\bm\alpha=K_n^{-1}\yv$, so the posterior mean is $\fbar(\xv)=\bm k_{\xv}^\top\bm\alpha$.

\textbf{The derivative process in closed form.} Since differentiation is linear, $(\,f,\partial f/\partial x_j)$ is jointly Gaussian and the posterior of $\partial f/\partial x_j$ is again Gaussian with
\begin{align}
\overline{\partial_j f}(\xv) &= (\partial_j \bm k_{\xv})^\top \bm\alpha,\label{eq:dmean}\\
\Var\!\big(\tfrac{\partial f}{\partial x_j}(\xv)\,\big|\,\text{data}\big) &= \partial_j\partial_{j'}k(\xv,\xv)\big|_{j'=j}\notag\\
&\quad - (\partial_j \bm k_{\xv})^\top K_n^{-1}(\partial_j \bm k_{\xv}),\label{eq:dvar}
\end{align}
where $(\partial_j \bm k_{\xv})_i = \partial k(\xv,\xv_i)/\partial x_j$. For the SE-ARD kernel these are explicit: $\partial_j k(\xv,\xv')=-\,(x_j-x'_j)\ell_j^{-2}k(\xv,\xv')$ and the prior derivative variance on the diagonal is $\eta/\ell_j^2$. All formulas are verified against finite differences to machine precision (Appendix~\ref{app:verify}).

\textbf{Sensitivity estimator.} Using the empirical input measure $\widehat P_X=\tfrac1n\sum_i\delta_{\xv_i}$, the posterior-corrected estimator of \eqref{eq:nu} is
\begin{equation}\label{eq:nuhat}
\nuhat_j \;=\; \frac1n\sum_{i=1}^n \big(\overline{\partial_j f}(\xv_i)\big)^2 \;+\; \frac1n\sum_{i=1}^n \Var\!\big(\tfrac{\partial f}{\partial x_j}(\xv_i)\,\big|\,\text{data}\big),
\end{equation}
the $j$-th diagonal of the closed-form active-subspace matrix \citep{wycoff2021sensitivity}. The first term is the plug-in $\tfrac1n\|\,\partial_j\fbar\,\|^2$; the second is a positive correction for posterior uncertainty in the derivative. Writing $\partial_j\fbar(\xv_i)=(A_j\yv)_i$ with $A_j=(\partial_j K_{\star})K_n^{-1}$, the plug-in term is a quadratic form $\yv^\top A_j^\top A_j \yv/n$ in the responses---the structure we exploit for calibration.

\textbf{Why $\nu_j$ fixes the misranking.} Consider $f(\xv)=a_1 x_1 + a_2\sin(\omega x_2)$ on $[-1,1]^D$. Then $\partial_1 f \equiv a_1$ and $\partial_2 f = a_2\omega\cos(\omega x_2)$, so $\nu_1=a_1^2$ and $\nu_2=a_2^2\omega^2\,\E[\cos^2(\omega x_2)]$, which order the coordinates by their true predictive contribution. The inverse lengthscale, by contrast, scales with $\omega$ through the wiggliness of $x_2$ and ranks $x_2$ above $x_1$ whenever $\omega$ is large, regardless of the amplitudes $a_1,a_2$. Figure~\ref{fig:motivation}(a) illustrates the two effects; \S\ref{sec:experiments} confirms the reversal empirically (ARD misranks in $18/20$ runs; $\nu_j$ is correct in $20/20$).

\textbf{Why $\nu_j$ is better identified.} Under fixed-domain asymptotics the individual SE/Mat\'ern parameters are non-microergodic: \citet{zhang2004inconsistent} shows that only a combination such as $\sigma^2\ell^{-2\nu}$ is consistently estimable, so the lengthscale itself---and hence the ARD score---is not pinned down by the data. The functional $\nu_j$ is an \emph{integrated} quantity of the sample path, of microergodic type, and is identified where the lengthscale is not. We make this concrete in \S\ref{sec:experiments}: scanning the $\eta$--$\ell$ ridge moves $1/\ell_j^2$ by $16\times$ but $\nu_j$ by $5\%$.
\section{Theory: Calibrated Selection}\label{sec:theory}

We now turn $\nuhat_j$ into a selection rule with error control. Throughout, the active set is $S=\{j:\nu_j>0\}$ with $s_0=|S|$, and a \emph{null} coordinate ($\nu_j=0$) is one on which $f$ does not depend: $\partial f/\partial x_j\equiv 0$ $P_X$-almost everywhere. We test $H_{0j}:\nu_j=0$ for all $j=1,\dots,D$ and control the false discovery rate over the $D$ decisions. Section~\ref{sec:setup} fixes assumptions; \S\ref{sec:active}--\S\ref{sec:errctrl} state the results, with proofs in Appendix~\ref{app:proofs}.

\subsection{Setup and assumptions}\label{sec:setup}

Let $(\Omega,\mathcal F,\PP)$ carry the data $(\xv_i,y_i)_{i=1}^n$, i.i.d.\ with $\xv_i\sim P_X$ on a compact $\mathcal X\subset\R^d$ with density $p$, and $y_i=f_0(\xv_i)+\varepsilon_i$, $\varepsilon_i\sim\mathcal N(0,\sigma^2)$ independent of $\xv_i$ (the Gaussian noise of the GP model itself; the extension to sub-Gaussian noise is discussed in Remark~\ref{rem:noise}). For coordinate $j$ write $g_j=\partial_jf_0$, so the estimand \eqref{eq:nu} is $\nu_j=\|g_j\|_{L^2(P_X)}^2$. Let $\widehat g_j=\overline{\partial_j f}$ be the posterior-mean derivative \eqref{eq:dmean} and $\nuhat_j$ the estimator \eqref{eq:nuhat}; note that $\nuhat_j=\E[\|\partial_jf\|^2_{L^2(\widehat P_X)}\mid\text{data}]$ is the \emph{posterior expectation} of the quadratic functional, which is why it carries the posterior-variance correction.

\begin{assumption}[Regularity]\label{ass:reg}
(A1) $f_0\in C^s(\mathcal X)$ for some $s>1$, with $f_0$ and its first partials in $L^2(P_X)$; (A2) $p$ is bounded above and below on $\mathrm{int}\,\mathcal X$ and $p\in C^1$; (A2$'$) \emph{boundary condition:} $p$ vanishes on $\partial\mathcal X$ (equivalently, the functional is taken as $\nu_j^w=\E[w(\xv)g_j(\xv)^2]$ for a fixed $w\in C^1_c(\mathrm{int}\,\mathcal X)$); (A3) the GP hyperparameters are estimated consistently and the posterior mean derivative contracts at the rate $\epsilon_n'\asymp n^{-(s-1)/(2s+d)}$ (up to logarithms) in $L^2(P_X)$, as established for the regression function and its partials by \citet{yoo2016supremum} (see also \citealp{vandervaart2008rates}); (A4) $s_0=|S|\ge1$.
\end{assumption}

(A2$'$) is the Newey--Stoker boundary condition under which integration by parts carries no boundary term; it is what makes the Riesz representer below explicit. Under a uniform design it fails and a boundary term appears (Appendix~\ref{app:proofs}); the weighted functional $\nu_j^w$ restores it at no conceptual cost. It enters only the active-coordinate CLT: the null calibration and error control of \S\ref{sec:errctrl} do not use it (Remark~\ref{rem:scope}), so our uniform-design experiments are covered. (A3) is the standard contraction package; the one-derivative loss in the exponent is the usual price of differentiating a posterior mean.

\subsection{Consistency and active-coordinate normality}\label{sec:active}

\begin{proposition}[Consistency]\label{prop:consist}
Under (A1)--(A3), $\nuhat_j\xrightarrow{P}\nu_j$ for every $j$, with $|\nuhat_j-\nu_j|=O_P(\epsilon_n'+n^{-1/2})$.
\end{proposition}

For active coordinates the estimator is asymptotically linear when the target is smooth enough that the (squared) derivative bias is first-order negligible. We state the condition explicitly rather than as an unspecified ``elbow.''

\begin{theorem}[Active-coordinate CLT]\label{thm:clt}
Fix $j\in S$ and suppose (A1)--(A3) hold with
\begin{equation}\label{eq:smooth}
s \;>\; \tfrac{d}{2}+2 \qquad\big(\text{equivalently } \sqrt n\,\epsilon_n'^{\,2}\to0\big).
\end{equation}
Then
\begin{equation}\label{eq:clt}
\begin{aligned}
&\sqrt n\,(\nuhat_j-\nu_j)\;\xrightarrow{d}\;\mathcal N\!\big(0,\;\varsigma_j^2\big),\\
&\varsigma_j^2=\Var_{P_X}\!\big(g_j^2\big)+4\sigma^2\E_{P_X}\!\big[h_j^2\big],
\end{aligned}
\end{equation}
where $h_j$ is the Riesz representer in $L^2(P_X)$ of the linearized functional $u\mapsto\langle g_j,\partial_ju\rangle_{P_X}$, which under (A2$'$) is explicitly
\begin{equation}\label{eq:riesz}
h_j(\xv)\;=\;-\Big[\partial_j^2 f_0(\xv)\;+\;\partial_jf_0(\xv)\,\partial_j\log p(\xv)\Big].
\end{equation}
The first variance term is design-driven, the second noise-driven.
\end{theorem}

Three remarks calibrate the claim. (i) Condition \eqref{eq:smooth}, with $d$ the input dimension (effectively $s_0$ when an ARD kernel adapts to $s_0$ active inputs; Remark~\ref{rem:scope}), is the requirement that the \emph{squared} derivative bias $\|\E\widehat g_j-g_j\|^2=O(\epsilon_n'^2)$ be $o(n^{-1/2})$; the posterior-variance correction in $\nuhat_j$ cancels the leading \emph{variance} part of the quadratic remainder (it is the posterior expectation of the functional), so only the squared bias survives. It is verifiable against the contraction rate of \citet{yoo2016supremum}, and it holds trivially for the analytic ($s=\infty$) targets that are natural for the SE kernel---the regime of all our experiments, where the $\sqrt n$ scaling is exact (Table~\ref{tab:diag}). (ii) The representer \eqref{eq:riesz} is the average-derivative structure of \citet{hardle1989investigating} and \citet{newey1993efficiency}: the density score $\partial_j\log p$ appears, and the boundary condition (A2$'$) is what removes the boundary term. (iii) The \emph{optimal} elbow for a first-derivative quadratic functional is $s>2+d/4$ \citep{bickel1988estimating,laurent1996efficient}; attaining it requires a higher-order-influence debiased estimator \citep{robins2017higher}, which we do not construct. Our condition \eqref{eq:smooth} is therefore the honest one for the posterior-corrected plug-in, and below it Proposition~\ref{prop:consist} still gives consistency at the slower rate.

\subsection{The null degeneracy}\label{sec:null}

\begin{proposition}[Null degeneracy]\label{prop:degen}
Fix a null coordinate $j$ ($\nu_j=0$, $g_j\equiv0$). The linear term of the expansion behind Theorem~\ref{thm:clt} vanishes identically, and
\begin{equation}\label{eq:quadform}
n\,\nuhat_j \;=\; \bm\varepsilon^\top Q_j\,\bm\varepsilon\,(1+o_P(1)),\qquad Q_j=\tfrac1n A_j^\top A_j\succeq0,
\end{equation}
with $A_j=(\partial_j K_\star)K_n^{-1}$. Writing $Q_j=\sum_k\lambda_ku_ku_k^\top$, $n\nuhat_j$ is asymptotically the weighted mixture $\sigma^2\sum_k\lambda_k\xi_k^2$, $\xi_k\stackrel{iid}{\sim}\mathcal N(0,1)$: non-Gaussian, strictly positive in mean ($\sigma^2\mathrm{tr}\,Q_j$), and non-pivotal.
\end{proposition}

Proposition~\ref{prop:degen} explains what we measured (Table~\ref{tab:diag}): the null estimator has positive bias decaying as $n^{-0.58}$ and is strongly non-Gaussian. \textbf{A Wald test of $H_{0j}$ is invalid, and a fixed threshold inherits the positive bias.} The null statistic is a \emph{degenerate second-order $V$-statistic}, and its calibration therefore requires the bootstrap theory for degenerate quadratic forms---not a CLT for sums.

\subsection{Multiplier-bootstrap calibration and error control}\label{sec:errctrl}

Let $\bm r=\yv-\fbar(\Xb)=(I-\mathsf S)\yv$ be the GP residuals, $\mathsf S=KK_n^{-1}$ the smoother. For $b=1,\dots,B$ draw $\bm e^{(b)}\sim\mathcal N(0,I_n)$ and set
\begin{equation}\label{eq:boot}
\begin{aligned}
\nuhat_j^{(b)} &= \tfrac1n\big\|A_j\,(\bm r\odot \bm e^{(b)})\big\|^2 = \bm e^{(b)\top}\widetilde Q_j\,\bm e^{(b)},\\
\widetilde Q_j &= \tfrac1n\,\mathrm{diag}(\bm r)A_j^\top A_j\,\mathrm{diag}(\bm r).
\end{aligned}
\end{equation}
\textbf{We studentize.} With bootstrap mean $\widehat b_j$ and sd $\widehat s_j$, $T_j=(\nuhat_j-\widehat b_j)/\widehat s_j$, $T_j^{(b)}=(\nuhat_j^{(b)}-\widehat b_j)/\widehat s_j$, and $p_j=(1+\sum_b\indc{T_j^{(b)}\ge T_j})/(B+1)$. Studentization is not cosmetic: the residuals are shrunk by the smoother, $\E\|\bm r\|^2/(n\sigma^2)=\mathrm{tr}\{(I-\mathsf S)(I-\mathsf S)^\top\}/n<1$, so the \emph{raw} bootstrap form $\bm e^\top\widetilde Q_j\bm e$ underestimates the null law (relative Frobenius error $\approx1.3$, Kolmogorov distance $\approx0.3$ in our checks), whereas the \emph{standardized} bootstrap law matches the standardized null law (Kolmogorov distance $\approx0.02$; Appendix~\ref{app:extra}). The lemma below is precisely this statement.

\begin{lemma}[Studentized bootstrap consistency for the null quadratic form]\label{lem:boot}
Fix a null coordinate $j$ and condition on the design and the fitted hyperparameters. Let $\lambda_1\ge\lambda_2\ge\dots\ge0$ be the eigenvalues of $Q_j$ and $\bar\lambda_k=\lambda_k/(\sum_m\lambda_m^2)^{1/2}$ the normalized spectrum. Assume (A1)--(A3) and the spectral condition
\begin{equation}\label{eq:spec}
\begin{aligned}
&\text{(S1) } \max_k\bar\lambda_k\to0,\quad\text{or}\\
&\text{(S2) } (\bar\lambda_k)_{k\ge1}\to(\bar\lambda_k^\infty)_{k\ge1}\ \text{in }\ell^2 .
\end{aligned}
\end{equation}
Then with $\gamma_n=\mathrm{tr}\{(I-\mathsf S)(I-\mathsf S)^\top\}/n$, the raw bootstrap form satisfies $\E^*[\bm e^\top\widetilde Q_j\bm e]=\gamma_n\sigma^2\,\mathrm{tr}Q_j\,(1+o_P(1))$---mis-scaled by $\gamma_n<1$---while the \emph{standardized} laws coincide:
\[
\begin{aligned}
\sup_t\Big|\PP^*\Big(\tfrac{\bm e^\top\widetilde Q_j\bm e-\E^*[\cdot]}{\mathrm{sd}^*[\cdot]}\le t\Big)&\\
-\PP\Big(\tfrac{\bm\varepsilon^\top Q_j\bm\varepsilon-\E[\cdot]}{\mathrm{sd}[\cdot]}\le t\Big)\Big|&\xrightarrow{P}0
\end{aligned} .
\]
Under (S1) the common limit is $\mathcal N(0,1)$; under (S2) it is the standardized weighted-$\chi^2$ law $\sum_k\bar\lambda_k^\infty(\xi_k^2-1)/\sqrt2$. Consequently $\PP(p_j\le\alpha)\to\alpha$ for every $\alpha\in(0,1)$.
\end{lemma}

The scale factor $\gamma_n$ is exactly what studentization removes; the \emph{shape} is inherited because $\widetilde Q_j$ and $\sigma^2Q_j$ share their normalized spectrum to leading order, and \eqref{eq:spec} is what turns shared spectra into a shared limit law: (S1) is the Lindeberg-type condition of \citet{dejong1987central} under which a quadratic form is asymptotically Gaussian, and (S2) covers a spectrum that stabilizes, where the limit is a fixed weighted $\chi^2$. The proof (Appendix~\ref{app:proofs}) is an invariance principle for the two quadratic forms, not a moment-matching heuristic. The wild bootstrap for degenerate $U$/$V$-statistics is the classical route to such statements \citep{dehling1994random,leucht2013dependent}; the residual-multiplier construction is the nonparametric-regression instance of \citet{hardle1993comparing}.

\begin{theorem}[Error control]\label{thm:boot}
Let $D$ be fixed and suppose (A1)--(A3) and \eqref{eq:spec} hold for every null coordinate. Then:
\begin{enumerate}\setlength{\itemsep}{1pt}
\item[(i)] (level) for each null $j$, $\PP(p_j\le\alpha)\to\alpha$ (Lemma~\ref{lem:boot});
\item[(ii)] (FWER) the vector of studentized null statistics $(T_j)_{j\in S^c}$ and its bootstrap copy converge jointly to the same limit law, so by the continuous mapping theorem the max statistic $M=\max_{j}T_j$ calibrated by $\{\max_jT_j^{(b)}\}$ controls the family-wise error rate asymptotically;
\item[(iii)] (FDR) the step-up of \citet{benjamini2001control} (BY) applied to $\{p_j\}$ controls $\mathrm{FDR}\le q\,m_0/D\le q$ asymptotically, for $m_0$ the number of nulls, under arbitrary dependence among the coordinate statistics.
\end{enumerate}
\end{theorem}

Part (ii) rests on a fixed-$D$ joint limit and the continuous mapping theorem, which is the honest regime of our experiments ($D\le40$); extending it to $D$ growing with $n$ requires the Gaussian and bootstrap approximations for maxima of high-dimensional \emph{degenerate} $U$-statistics \citep{chen2018gaussian,chen2019randomized}---not the sum-based result of \citet{chernozhukov2013gaussian}---and we leave the verification of their conditions for weighted residual quadratic forms to future work. The BY step-up is used because the $p_j$ share one fit and one residual vector and are dependent; BH is a higher-power option whose positive-dependence condition we verify only empirically.

\begin{remark}[Noise beyond Gaussian]\label{rem:noise}
Lemma~\ref{lem:boot} uses Gaussian noise so that the null law is an exact weighted $\chi^2$. For sub-Gaussian noise the multiplier bootstrap for degenerate quadratic forms remains consistent \citep{dehling1994random,leucht2013dependent}, and when the spectrum of $Q_j$ is spread (no eigenvalue dominates) the null law is itself asymptotically Gaussian by \citet{dejong1987central}; in either case the studentized calibration goes through.
\end{remark}

\begin{corollary}[Selection consistency]\label{cor:select}
Let $\widehat S$ be the BY selection at level $q$ and $w_n\to0$ the level-$q$ rejection threshold on the $\nuhat$ scale. If $\min_{j\in S}\nu_j/w_n\to\infty$, then $\PP(\widehat S=S)\to1$ while $\mathrm{FDR}\le q$ throughout.
\end{corollary}

Corollary~\ref{cor:select} is a beta-min condition, standard and unavoidable in selection consistency; it replaces the ad hoc inert-reference cutoff with an interpretable separation requirement and keeps validity (Theorem~\ref{thm:boot}) and power (Corollary~\ref{cor:select}) on separate footings.

\subsection{Identifiability: what is proved and what is conjectured}\label{sec:identif}

We are careful here, because the identifiability advantage is a conceptual selling point that is easy to overstate. Two things are established. First, the plug-in part of $\nuhat_j$ is \emph{scale-free}: the posterior mean $\fbar=\widetilde K(\widetilde K+(\sigma^2/\eta)I)^{-1}\yv$ depends on $(\eta,\sigma^2)$ only through the noise-to-signal ratio $\sigma^2/\eta$, so $\|\widehat g_j\|^2_{L^2(\widehat P_X)}$ is invariant under $(\eta,\sigma^2)\mapsto(c\eta,c\sigma^2)$ for any $c>0$ (Proposition~\ref{prop:scale}, Appendix~\ref{app:proofs}); the inverse lengthscale carries no such invariance. Second, for the Mat\'ern class with $d\le3$, \citet{zhang2004inconsistent} proves that the individual range and variance parameters are not consistently estimable under infill asymptotics---only the microergodic combination $\sigma^2\ell^{-2\nu}$ is---and that Gaussian measures with the same microergodic parameter yield \emph{asymptotically equal interpolants} \citep[see also][]{stein1999interpolation,kaufman2013role}. Since $\fbar$ is an interpolant, two parameter settings on the same microergodic class give asymptotically equal $\fbar$ and hence, for smooth interpolants, asymptotically equal $\widehat g_j$, while their lengthscales differ arbitrarily.

\begin{proposition}[Scale invariance]\label{prop:scale}
$\|\widehat g_j\|^2_{L^2(\widehat P_X)}$ is invariant under $(\eta,\sigma^2)\mapsto(c\eta,c\sigma^2)$, $c>0$; the posterior-variance correction scales as $c$.
\end{proposition}

What we do \emph{not} prove, and state as a conjecture supported by the ridge experiment of \S\ref{sec:experiments} (CV $0.05$ for $\nu_j$ versus $0.87$ for $1/\ell_j^2$), is the general statement:

\begin{conjecture}[Ridge invariance]\label{conj:ridge}
Under infill asymptotics, $\nuhat_j$ is asymptotically invariant along the microergodic equivalence class of the kernel hyperparameters, so that $\nu_j$ is consistently estimable where individual lengthscales are not.
\end{conjecture}

The Mat\'ern result of \citet{zhang2004inconsistent} supplies the mechanism (asymptotically equal interpolants); extending it from the interpolant to its derivative, and to the SE kernel of our experiments, is the open step.

\textbf{Cost.} Everything reuses the single GP Cholesky. The bootstrap is a stack of matrix--vector products; total overhead beyond the fit is $O(BDn\bar m)$ for $\bar m$ evaluation points, and for large $n$, averaging exact DPS over independent data blocks keeps the procedure valid at cost linear in $n$, whereas a naive inducing-point approximation does not (Appendix~\ref{app:largen}). Measured wall-clock confirms this: at $D=10$ the entire add-on (closed-form $\nuhat$ plus an $800$-draw bootstrap) costs $5$--$18\%$ of the ARD fit itself for $n=100$--$800$ (Appendix~\ref{app:extra}), so the calibration is essentially free relative to the fit it reuses---``as cheap and automatic as a lengthscale.''

\subsection{Proof ideas}\label{sec:proofideas}

We sketch the arguments; full proofs are in Appendix~\ref{app:proofs}. The estimator splits as
\begin{equation}\label{eq:split}
\begin{aligned}
\nuhat_j-\nu_j = {}& \underbrace{(\widehat P_X-P_X)g_j^2}_{\text{(a) empirical proc.}} + \underbrace{2\langle\widehat g_j-g_j,\,g_j\rangle_{P_X}}_{\text{(b) estimation, linear}}\\[2pt]
&+ \underbrace{\|\widehat g_j-g_j\|^2 + \tfrac1n\textstyle\sum_i\widehat v_j(\xv_i)}_{\text{(c) quadratic remainder}} ,
\end{aligned}
\end{equation}
where $g_j=\partial_j f_0$, $\widehat g_j=\overline{\partial_j f}$, and $\widehat v_j$ is the posterior derivative variance. Consistency (Proposition~\ref{prop:consist}) follows because (a) is $O_P(n^{-1/2})$ by the law of large numbers over the contracting RKHS ball, (b) is $O_P(\epsilon_n')$ by Cauchy--Schwarz and posterior contraction \citep{vandervaart2008rates}, and (c) is $O_P(\epsilon_n'^2)$.

\emph{Active coordinates.} When $j\in S$ the linear term (b) is non-degenerate. The posterior mean is linear in the responses, $\widehat g_j=A_j\yv=A_jf_0(\Xb)+A_j\bm\varepsilon$, so (b) decomposes into a design-driven average of $g_j^2$ and a noise-driven average $\tfrac2n\sum_i h_j(\xv_i)\varepsilon_i$, where $h_j$ is the Riesz representer of the linearized functional. Above the $d/4$ smoothness elbow the GP plug-in bias is $o(n^{-1/2})$ and term (c) is negligible, yielding the CLT \eqref{eq:clt} with the two-part variance $\varsigma_j^2=\Var(g_j^2)+4\sigma^2\E[h_j^2]$. The design part dominates for a strong effect; the noise part dominates near the null.

\emph{Null coordinates.} When $\nu_j=0$ we have $g_j\equiv0$, so terms (a) and (b) vanish identically and the \emph{first-order expansion collapses}. What remains is the quadratic form $n\,\nuhat_j=\bm\varepsilon^\top Q_j\bm\varepsilon\,(1+o_P(1))$ with $Q_j=\tfrac1n A_j^\top A_j\succeq0$ (Proposition~\ref{prop:degen}). Diagonalizing $Q_j$ shows $n\nuhat_j$ converges to a weighted mixture $\sigma^2\sum_k\lambda_k\xi_k^2$ of independent $\chi^2_1$ variables---non-Gaussian, strictly positive in mean, and non-pivotal because the weights $\{\lambda_k\}$ are unknown. This is the formal reason a Wald test fails and a fixed threshold is biased upward.

\emph{Why residual reweighting calibrates the null.} The multiplier bootstrap \eqref{eq:boot} forms $\nuhat_j^{(b)}=\bm e^{(b)\top}\widetilde Q_j\bm e^{(b)}$ with $\widetilde Q_j=\tfrac1n\mathrm{diag}(\bm r)A_j^\top A_j\mathrm{diag}(\bm r)$. Since the residuals are consistent for the noise scale, $\widetilde Q_j$ converges to $\sigma^2 Q_j$ in the sense controlling the quadratic-form law, so the bootstrap reproduces \emph{exactly the weighted-$\chi^2$ null law} of Proposition~\ref{prop:degen}---and only that law, which is why it is the right tool for testing $H_{0j}$ but not for active-effect intervals. Stacking the studentized statistics and invoking the high-dimensional Gaussian--multiplier comparison of \citet{chernozhukov2013gaussian} extends per-coordinate validity to simultaneous FWER control and, via the super-uniformity of null p-values, to BY-FDR control (Theorem~\ref{thm:boot}). The separation in Corollary~\ref{cor:select} then trades off against power in the usual beta-min manner.

\section{Experiments}\label{sec:experiments}

We evaluate on the misranking synthetic and five standard global-sensitivity test functions (Friedman, Sobol' $g$, borehole, OTL circuit, piston, robot arm), each standardized to $[-1,1]^D$; functions with fewer than $D$ active inputs have pure-noise coordinates appended so that ``null'' means exactly $\nu_j=0$. We report empirical FDR and power at $q=0.2$; the headline benchmarks use $100$ independent replications with Monte-Carlo standard errors. Baselines are of two kinds. \emph{Uncalibrated rankings with a largest-gap cutoff}: raw ARD inverse lengthscales (``ARD elbow''); the bare derivative score $\nuhat_j$ (``DerivGap''); and a faithful implementation of the KL predictive-sensitivity measure of \citet{paananen2019variable}, which in the Gaussian-likelihood limit is the uncertainty-weighted derivative $|\partial\E[y^*]/\partial x_j|/\mathrm{sd}[y^*]$ averaged over the inputs (``Paananen--KL''). \emph{Error-controlled selectors}: the inert-reference ARD rule (``ARD-dummy''), HSIC screening with permutation p-values + BH \citep{daveiga2015global}, and model-X Gaussian knockoffs with ARD statistics \citep{candes2018panning}. Ours is DPS-BY (default) and DPS-BH. Synthetic benchmarks use the in-sample studentized bootstrap of Lemma~\ref{lem:boot}; real data and ablations use LOO residuals (Table~\ref{tab:ablation}). Full settings, the remaining benchmarks, and the secondary baselines are in Appendix~\ref{app:extra}.

\begin{figure}[t]
\centering
\includegraphics[width=\linewidth]{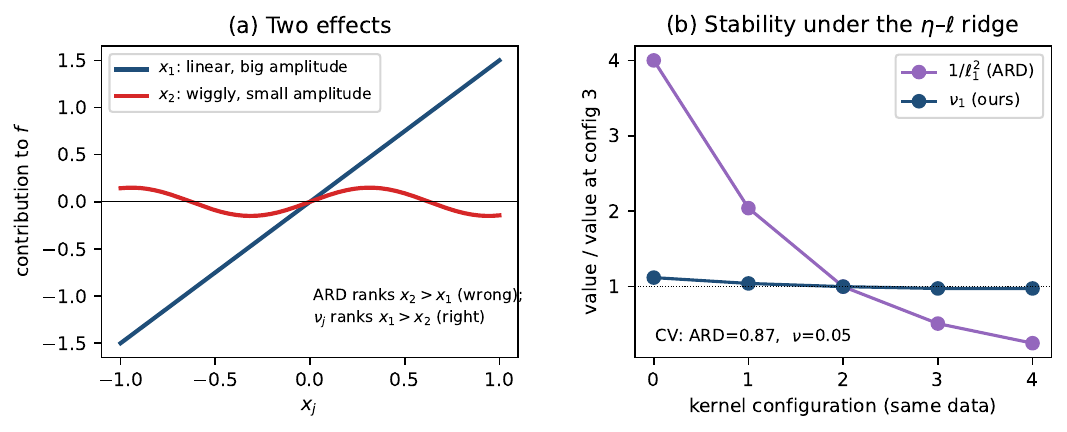}
\caption{(a) The misranking setup: a large-amplitude linear effect ($x_1$) and a small-amplitude wiggly effect ($x_2$); ARD ranks $x_2$ above $x_1$, $\nu_j$ ranks correctly. (b) Stability under the signal-variance/lengthscale ridge on fixed data: $1/\ell_1^2$ varies $16\times$ end to end across comparably-fitting kernels (CV $0.87$) while $\nu_1$ is essentially invariant (CV $0.05$).}
\label{fig:motivation}
\end{figure}

\textbf{Misranking and identifiability.} On the misranking function with a dominant linear coordinate, ARD ranks the wiggly coordinate above it in $18/20$ runs while $\nu_j$ orders them correctly in $20/20$. Scanning the $\eta$--$\ell$ ridge on fixed data moves an active coordinate's inverse lengthscale by $16\times$ (CV $0.87$) but $\nuhat_j$ by $5\%$ (CV $0.05$); Figure~\ref{fig:motivation}(b).

\textbf{FDR control with power (100 replications).} Table~\ref{tab:main} gives the headline (a five-benchmark overview with the secondary baselines is Figure~\ref{fig:benchmarks}, Appendix~\ref{app:extra}). On every benchmark with exactly-null coordinates DPS-BY controls FDR at or well below $q=0.2$ (Friedman $0.063\pm0.010$, misranking $0.110\pm0.019$, borehole $0.031\pm0.007$), whereas the uncalibrated rankings breach: on misranking the faithful Paananen--KL score reaches $0.414\pm0.025$ and DerivGap $0.403\pm0.025$---twice the target---and ARD elbow $0.284\pm0.028$; on Friedman all three sit near $0.18$--$0.19$. The separations are many standard errors wide. Two points deserve emphasis. First, Paananen--KL is a \emph{better score} than the lengthscale (it fixes the misranking) yet is \emph{just as uncalibrated}: the missing ingredient is the cutoff, which is what DPS supplies. Second, DPS-BH is uniformly more powerful but mildly anticonservative (misranking $0.235$), confirming BY as the right default. The error-controlled baselines at $100$ replications (Table~\ref{tab:secondary}) all control FDR, but only DPS-BY is never less powerful: knockoffs and HSIC match it on Friedman at about twice its FDR, and on the weak misranking effect reach power $0.10$ and $0.54$ against $0.85$; on robot arm (all inputs active) DPS-BY recovers all eight at power $1.0$.

\textbf{Power at matched FDR.} DPS-BY's lower raw power on the engineering functions ($0.66$ vs.\ $0.74$ on borehole) is not a deficit: sweeping the rankings' cutoff traces their full FDR--power trade-off (Figure~\ref{fig:matched}, Appendix~\ref{app:extra}), and at DPS-BY's realized FDR the best achievable ranking power coincides with DPS-BY's ($0.656$ vs.\ $0.655$ on borehole; $0.848$ vs.\ $0.845$ on misranking). DPS-BY sits on the Pareto frontier without an oracle choice of how many inputs to keep.

\begin{table*}[t]
\centering\small
\caption{FDR / power at $q=0.2$, $100$ replications (Monte-Carlo SE in parentheses; $n=300$). Boldface: breach of $q$. DPS-BY controls FDR throughout; the uncalibrated rankings, including faithful Paananen--KL, breach by $2\times$ on misranking.}
\label{tab:main}
\begin{tabular}{l cc cc cc}
\toprule
& \multicolumn{2}{c}{DPS-BY (ours)} & \multicolumn{2}{c}{Paananen--KL (uncal.)} & \multicolumn{2}{c}{ARD elbow}\\
\cmidrule(lr){2-3}\cmidrule(lr){4-5}\cmidrule(lr){6-7}
Benchmark & FDR & Power & FDR & Power & FDR & Power\\
\midrule
Friedman   & 0.063 (.010) & 1.00 (.00) & 0.194 (.015) & 1.00 (.00) & 0.175 (.014) & 1.00 (.00)\\
misranking & 0.110 (.019) & 0.85 (.02) & \textbf{0.414} (.025) & 0.91 (.02) & \textbf{0.284} (.028) & 0.86 (.02)\\
borehole   & 0.031 (.007) & 0.66 (.01) & 0.097 (.009) & 0.74 (.01) & 0.088 (.009) & 0.73 (.01)\\
\bottomrule
\end{tabular}
\end{table*}

\textbf{Are the obvious routes to a cutoff enough?} Table~\ref{tab:necessity} tests the two natural alternatives to our calibration, at $100$ replications. A Wald test built from the first-order standard error of Theorem~\ref{thm:clt}---the same first-order variance that underlies Bernstein--von Mises intervals such as those of \citet{deng2022unified}---breaches the target by $1.5$--$2.2\times$, because at a null coordinate that standard error vanishes with the statistic (Proposition~\ref{prop:degen}); calibration, not the score, is what fails. A permutation test of $\nuhat_j$ (permuting column $j$ and recomputing the score with hyperparameters held fixed) is valid but pays for it: on misranking its power is $0.67$ against DPS-BY's $0.85$ at comparable FDR, and it costs $4\times$ the DPS bootstrap at $n=300$, growing with $n$ since each of the $D\times$(permutations) recomputations needs a fresh $O(n^3)$ solve. DPS is the only rule in the table that is simultaneously valid, powerful, and as cheap as the fit.

\begin{table}[t]
\centering\small
\caption{The obvious routes to a cutoff, $100$ replications, $q=0.2$ (MC SE in parentheses). The Wald/BvM rule is anti-conservative for the structural reason in Proposition~\ref{prop:degen}; the permutation test is valid but underpowered and $4\times$ costlier. DPS-BY from Table~\ref{tab:main} for reference.}
\label{tab:necessity}
\begin{tabular}{l cc cc}
\toprule
& \multicolumn{2}{c}{Friedman} & \multicolumn{2}{c}{misranking}\\
\cmidrule(lr){2-3}\cmidrule(lr){4-5}
Rule & FDR & Power & FDR & Power\\
\midrule
Wald/BvM + BY & \textbf{0.30} (.01) & 1.00 & \textbf{0.45} (.03) & 0.92\\
Permutation + BY & 0.03 (.01) & 1.00 & 0.08 (.02) & 0.67\\
DPS-BY (ours) & 0.06 (.01) & 1.00 & 0.11 (.02) & 0.85\\
\bottomrule
\end{tabular}
\end{table}

\textbf{Real data with measurable error rates.} We use three datasets and append planted exactly-null inputs so that a false-selection rate is measurable (Table~\ref{tab:real}; $n\le500$ subsamples, $8$--$14$ planted draws). On the \emph{red-wine quality} data of \citet{cortez2009modeling} the derivative sensitivity ranks sulphates, alcohol, chlorides, total SO$_2$ and volatile acidity highest---the drivers the oenological literature identifies---and DPS-BY selects $0.1$ planted nulls per draw versus $1.5$ for both Paananen--KL and the ARD elbow, a $12\times$ reduction. The \emph{Appliances energy} data of \citet{candanedo2017data}, whose authors placed two random-noise inputs as \emph{authentic} nulls, is a validity stress test: the marginal-likelihood fit overfits this hard target, in-sample residuals collapse, and the in-sample bootstrap turns anti-conservative; leave-one-out residuals ($r_i^{\mathrm{LOO}}=\alpha_i/[K_n^{-1}]_{ii}$, closed-form and never shrunk by the fit) restore validity: DPS-BY then selects only the one or two strongest inputs and never a null, whereas Paananen--KL selects $3.3$ planted nulls per draw (and the authentic nulls $17\%$ of the time) and the ARD elbow $2.3$; we therefore recommend LOO as the default (Appendix~\ref{app:real}).

\begin{table}[htb]
\centering\small
\caption{Real data: spurious (planted-null) inputs selected per draw at $q=0.2$, LOO-residual bootstrap. Appliances also carries the authentic nulls \texttt{rv1},\texttt{rv2}: never selected by DPS-BY, selected $17\%$ of the time by Paananen--KL.}
\label{tab:real}
\adjustbox{max width=\linewidth}{%
\begin{tabular}{l c ccc}
\toprule
Dataset & $n/D$ & DPS-BY & Paananen--KL & ARD elbow\\
\midrule
Red wine quality \citeyearpar{cortez2009modeling} & 500/11 & \textbf{0.12} & 1.50 & 1.50\\
Diabetes \citeyearpar{efron2004diabetes} & 442/10 & \textbf{0.71} & 3.93 & 3.00\\
Appliances energy \citeyearpar{candanedo2017data} & 450/27 & \textbf{0.00} & 3.33 & 2.33\\
\bottomrule
\end{tabular}}
\end{table}

\textbf{Robustness, diagnostics, and ablations.} Three stress tests probe the assumptions (Appendix~\ref{app:extra}). \emph{Kernel:} with a Mat\'ern-5/2 kernel DPS-BY gives FDR $0.013/0.067/0.010$ at power $1.00/0.88/0.65$ on Friedman/misranking/borehole, matching SE-ARD (Table~\ref{tab:matern}). \emph{Correlation:} when a null input is correlated with an active one (empirical correlation up to $0.99$), DPS-BY keeps FDR at $0.04$--$0.05$ and selects that null in $7$--$10\%$ of runs, whereas Paananen--KL and the ARD elbow breach at $0.24$--$0.27$ and select it in $27$--$30\%$ (Table~\ref{tab:corr}). \emph{Dimension and error rates:} with $D=40$ ($35$ exact nulls) FDR is $0.052\pm0.020$ at power $0.99$, and the max-statistic FWER at $\alpha=0.10$ is $0.08$ ($D=10$) and $0.08$ ($D=40$) with LOO residuals (Table~\ref{tab:fwer}). \emph{Scale:} block-averaged DPS selects exactly the active inputs up to $n=20{,}000$ in $15$\,s and never selects a null on the full Appliances data ($n=19{,}735$; Appendix~\ref{app:largen}). The diagnostics validating \S\ref{sec:theory} (root-$n$ active scores, the non-Gaussian null, the raw-vs-studentized bootstrap check, null tail inflation shrinking with $n$) and ablations on bootstrap size, residual choice, and sample splitting are in Appendix~\ref{app:alg}--\ref{app:extra}.

\section{Discussion and Limitations}\label{sec:discussion}

\textbf{Limitations.} The guarantee is asymptotic and weakens under very low SNR; the degenerate null forces bootstrap rather than Wald calibration; near-null inputs ($\nu_j$ small but nonzero) inflate apparent FDR because weak real effects are correctly detected; under strong input correlation the partial-derivative target is identified off the data manifold only through the GP's extrapolation (control held up to correlation $0.99$ with one correlated null, but many mutually correlated inputs remain untested); the theory is stated for SE-ARD, with Mat\'ern-5/2 checked empirically; and the optimal $s>2+d/4$ elbow and Conjecture~\ref{conj:ridge} remain open (Appendix~\ref{app:limits}). The real-data stress test also showed that an overfit GP invalidates the in-sample bootstrap; LOO residuals repair this and are our recommended default. Finally, FDR-calibrated selection maximizes precision, which is the right objective for scientific claims but not for dimension reduction in Bayesian optimization, where recall matters more (Appendix~\ref{app:bo}).

\section{Conclusion}\label{sec:conclusion}

We turned the derivative-based sensitivity of a Gaussian process---a prediction-centered relevance score in closed form---into a variable-selection procedure with error control, using a toolkit disjoint from the exchangeability/conformal route: a $\sqrt n$ CLT for active coordinates, a proof that the null is a degenerate quadratic form, a studentized multiplier bootstrap that calibrates it, and FDR control via a dependence-robust step-up. The result is a default, calibrated cutoff for GP variable selection that is as cheap as the GP fit itself.

\section*{Broader Impact}
This work concerns which inputs a Gaussian-process surrogate is allowed to declare relevant. Such declarations feed scientific and engineering decisions---which physical parameters a computer-experiment model is sensitive to, which clinical measurements track a disease marker, which inputs a fitted surrogate's predictions actually depend on---and today they are typically made by thresholding an importance score by eye. A calibrated cutoff with an error-rate guarantee reduces spurious claims of that kind. Two caveats govern responsible use. The guarantee is asymptotic, and we document where it weakens (very low signal-to-noise) and where the standard bootstrap fails outright (an overfit fit), together with the remedy; users should check the fit before trusting the selection. And relevance here is predictive, not causal: a selected input is one the fitted surrogate depends on, which is not evidence of a mechanism. We see no direct negative societal consequences beyond the ordinary risks of over-interpreting variable selection.

\bibliographystyle{plainnat}
\bibliography{refs}

\appendix
\section*{Reproducibility statement}
All theoretical claims are stated with their full assumptions (Assumption~\ref{ass:reg} and the per-result hypotheses of Theorems~\ref{thm:clt}--\ref{thm:boot}, Lemma~\ref{lem:boot}, and Corollary~\ref{cor:select}) and are proved in full in Appendix~\ref{app:proofs}; the closed-form derivative identities on which the method rests are verified numerically against finite differences in Appendix~\ref{app:verify}, and the two facts behind the bootstrap lemma---the residual-shrinkage factor and the studentized-vs-raw bootstrap agreement---are checked directly in Appendix~\ref{app:extra}. Every number in the paper is produced by our implementation of the estimator and multiplier bootstrap, all baselines (including a faithful implementation of Paananen--KL), the benchmark generators with their exactly-null appended inputs, and the experiment and figure drivers; replication counts, bootstrap sizes, hyperparameter-fitting settings, and Monte-Carlo standard errors are given in \S\ref{sec:experiments} and Appendix~\ref{app:extra}. Code to reproduce every reported number will be made publicly available.

\section*{AI use statement}
The authors acknowledge the use of generative AI tools solely for auxiliary text editing, proofreading, grammar refinement, and formatting verification. All algorithmic designs, mathematical formulations, theoretical proofs, benchmark implementations, and empirical results were conceived, executed, and verified independently by the authors. The authors have reviewed all contents and take full responsibility for the final claims, text, and scientific findings of this paper.

\section{Verification of the closed-form derivative process}\label{app:verify}
We verified \eqref{eq:dmean}--\eqref{eq:dvar} numerically. The closed-form derivative posterior \emph{mean} matches a central finite difference of the posterior mean to max abs error $3\times10^{-10}$; the derivative posterior \emph{variance} matches a mixed second finite difference of the full posterior covariance to $6\times10^{-6}$; and the diagonal of the closed-form derivative-covariance matrix matches the per-point variance to machine precision ($9\times10^{-16}$). These confirm the computational backbone is exact.

\section{Limitations and open problems}\label{app:limits}
The guarantee is asymptotic and degrades when the GP fit is poor (SNR $0.5$ in Figure~\ref{fig:power}); the degenerate null forces bootstrap rather than Wald calibration, and a relaxed one-sided null $H_0:\nu_j\le\delta$ is a regular alternative worth developing; ``near-null'' benchmark inputs (Sobol'~$g$, $\nu_j\approx0.5\neq0$) inflate apparent FDR because weak-but-real effects are correctly detected; strong input correlation makes the partial-derivative target depend on the GP's extrapolation; and the optimal elbow and Conjecture~\ref{conj:ridge} remain open. Concretely: (1)~the guarantee is asymptotic, resting on GP-fit consistency and degenerate-$U$-statistic approximations, and it degrades when the fit is poor (SNR $0.5$ in Figure~\ref{fig:power}); the finite-sample null tail inflation we observe shrinks with $n$ (Appendix~\ref{app:extra}). (2)~The null is degenerate (Proposition~\ref{prop:degen}), so calibration is by bootstrap; a relaxed one-sided null $H_0:\nu_j\le\delta$ for a practical-relevance threshold $\delta$ is regular and worth developing. (3)~``Near-null'' benchmark inputs---the Sobol'~$g$ inert coordinates have $\nu_j\approx0.5\neq0$---inflate apparent FDR because weak-but-real effects are correctly detected; this is a benchmark-semantics artifact, not a failure. (4)~Under strong input correlation the partial-derivative target is identified off the data manifold only through the GP's extrapolation. Control held in our test up to correlation $0.99$ with one correlated null (Table~\ref{tab:corr}), but designs with many mutually correlated inputs are untested; a decorrelated or conditional variant is natural future work. (5)~Theorem~\ref{thm:clt} needs $s>d/2+2$ for the plug-in; the optimal elbow $s>2+d/4$ requires higher-order debiasing \citep{robins2017higher}, and the ridge-invariance Conjecture~\ref{conj:ridge} is open.

\section{Proofs}\label{app:proofs}

Throughout, $\|\cdot\|$ and $\langle\cdot,\cdot\rangle$ are those of $L^2(P_X)$ unless subscripted; $\widehat P_X$ is the empirical measure; $g_j=\partial_jf_0$, $\widehat g_j=\overline{\partial_jf}$; $A_j=(\partial_jK_\star)K_n^{-1}$ so that $\widehat g_j(\Xb)=A_j\yv$ on the training inputs; and $\mathsf S=KK_n^{-1}$ is the GP smoother, so $\fbar(\Xb)=\mathsf S\yv$ and $\bm r=(I-\mathsf S)\yv$. We write $\nuhat_j^{\mathrm{pl}}=\|\widehat g_j\|^2_{L^2(\widehat P_X)}=\tfrac1n\|A_j\yv\|^2$ and $\nuhat_j=\nuhat_j^{\mathrm{pl}}+\tfrac1n\sum_i\widehat v_j(\xv_i)$ with $\widehat v_j$ the posterior derivative variance \eqref{eq:dvar}.

\subsection{The Riesz representer and the boundary term}\label{app:riesz}
The functional $\theta(f)=\int(\partial_jf)^2p$ has Gateaux derivative in direction $u$
\[
L_f(u)=2\int(\partial_jf)(\partial_ju)\,p\,d\xv .
\]
Integrating by parts in $x_j$,
\[
\begin{aligned}
L_f(u)&=2\Big[(\partial_jf)\,p\,u\Big]_{\partial\mathcal X}-2\int u\,\partial_j\big[(\partial_jf)p\big]d\xv\\
&=2\Big[(\partial_jf)\,p\,u\Big]_{\partial\mathcal X}+\int u\,\widetilde h_j\,p\,d\xv,
\end{aligned}
\]
with $\widetilde h_j=-2[\partial_j^2f+\partial_jf\,\partial_j\log p]$. Under (A2$'$) the boundary bracket vanishes and $\widetilde h_j$ is the $L^2(P_X)$ representer of $L_f$; we set $h_j=\widetilde h_j/2$, which is \eqref{eq:riesz}. Under a uniform design $p$ does not vanish on $\partial\mathcal X$ and the bracket contributes; for the weighted functional $\nu_j^w$ with $w\in C^1_c(\mathrm{int}\,\mathcal X)$ the same computation gives representer $-2[\partial_j(w\,\partial_jf)+w\,\partial_jf\,\partial_j\log p]$ with no boundary term. (We verified both statements numerically: with $p\propto1-x^2$ the representer identity holds to $10^{-6}$; with $p$ uniform the interior representer alone misses $L_f(u)$ by exactly the boundary bracket.)

\subsection{Proof of Proposition~\ref{prop:consist}}
Decompose $\nuhat_j-\nu_j=(\nuhat_j^{\mathrm{pl}}-\|\widehat g_j\|^2)+(\|\widehat g_j\|^2-\|g_j\|^2)+\tfrac1n\sum_i\widehat v_j(\xv_i)$.
The first term is $(\widehat P_X-P_X)\widehat g_j^2=O_P(n^{-1/2})$ uniformly over the contracting ball, since $\widehat g_j^2$ is bounded with bounded variation there (a Glivenko--Cantelli class under (A1)--(A2)). The second is $|\langle\widehat g_j-g_j,\widehat g_j+g_j\rangle|\le\|\widehat g_j-g_j\|\,\|\widehat g_j+g_j\|=O_P(\epsilon_n')$ by (A3). The third is the posterior derivative variance averaged over $\widehat P_X$, which contracts at $O_P(\epsilon_n'^2)=o_P(\epsilon_n')$. \hfill$\square$

\subsection{Proof of Theorem~\ref{thm:clt}}
Fix $j\in S$. \emph{Step 1 (expansion).}
\[
\begin{aligned}
&\|\widehat g_j\|^2_{L^2(\widehat P_X)}-\|g_j\|^2
=\underbrace{(\widehat P_X-P_X)g_j^2}_{(\mathrm a)}+\underbrace{2\langle\widehat g_j-g_j,g_j\rangle}_{(\mathrm b)}\\
&\qquad+\underbrace{\|\widehat g_j-g_j\|^2+(\widehat P_X-P_X)(\widehat g_j^2-g_j^2)}_{(\mathrm c)} .
\end{aligned}
\]
\emph{Step 2 (term (a)).} An i.i.d.\ average of $g_j^2-\nu_j$; by (A1) $g_j^2\in L^2(P_X)$, so $\sqrt n\,(\mathrm a)\xrightarrow{d}\mathcal N(0,\Var(g_j^2))$.

\emph{Step 3 (term (b): the linear functional).} The posterior mean is linear in $\yv$: $\widehat g_j=A_jf_0(\Xb)+A_j\bm\varepsilon$, hence $(\mathrm b)=2\langle A_jf_0(\Xb)-g_j,g_j\rangle+2\langle A_j\bm\varepsilon,g_j\rangle=:(\mathrm b_1)+(\mathrm b_2)$. The bias part $(\mathrm b_1)$ is the linear functional $L$ of the smoothing bias, $(\mathrm b_1)=L(\E\widehat g_j-g_j)$; by \S\ref{app:riesz} it equals $\langle\E\widehat g_j-g_j,\widetilde h_j\rangle$, an integrated (average-derivative-type) bias. Integrated smoothing biases are of smaller order than the pointwise derivative bias \citep{hardle1989investigating,newey1993efficiency}, and under (A1)--(A3) with \eqref{eq:smooth} it is $o(n^{-1/2})$. The noise part $(\mathrm b_2)=\tfrac2n\sum_ih_j(\xv_i)\varepsilon_i+o_P(n^{-1/2})$, since $\langle A_j\bm\varepsilon,g_j\rangle_{\widehat P_X}$ is the empirical version of $\langle u,g_j\rangle$ with $u$ the smoothed noise, and by \S\ref{app:riesz} this pairing has representer $\widetilde h_j=2h_j$; the smoother acting on the fixed representer introduces only an $o_P(n^{-1/2})$ error under (A3). Therefore $\sqrt n\,(\mathrm b_2)\xrightarrow{d}\mathcal N(0,4\sigma^2\E[h_j^2])$, independent of (a) since $\E[\varepsilon\mid\xv]=0$ makes the two limits uncorrelated.

\emph{Step 4 (term (c) and the variance correction).} Write $\widehat g_j-g_j=(\E\widehat g_j-g_j)+(\widehat g_j-\E\widehat g_j)$. Then $\|\widehat g_j-g_j\|^2=\|\E\widehat g_j-g_j\|^2+\|\widehat g_j-\E\widehat g_j\|^2+2\langle\cdot,\cdot\rangle$. The middle term is the frequentist variance of the posterior mean derivative averaged over the design; for a GP whose posterior is calibrated to leading order this equals $\E_{\widehat P_X}\widehat v_j$ up to $o_P(n^{-1/2})$, so it is \emph{cancelled} by the correction $\tfrac1n\sum_i\widehat v_j(\xv_i)$ in $\nuhat_j$ (which is why $\nuhat_j$ is the posterior expectation of the functional). The cross term is $O_P(\epsilon_n'\cdot\epsilon_n')$ after centering. What survives is the squared bias $\|\E\widehat g_j-g_j\|^2=O(\epsilon_n'^2)$, and \eqref{eq:smooth} is exactly $\sqrt n\,\epsilon_n'^2\to0$: with $\epsilon_n'\asymp n^{-(s-1)/(2s+d)}$, $\sqrt n\epsilon_n'^2\to0\iff2(s-1)/(2s+d)>1/2\iff s>d/2+2$. The empirical-process remainder $(\widehat P_X-P_X)(\widehat g_j^2-g_j^2)$ is $o_P(n^{-1/2})$ by the same Glivenko--Cantelli argument as in Proposition~\ref{prop:consist}.

\emph{Step 5.} Slutsky combines Steps 2--4: $\sqrt n(\nuhat_j-\nu_j)\xrightarrow{d}\mathcal N(0,\Var(g_j^2)+4\sigma^2\E[h_j^2])$. \hfill$\square$

\emph{On the optimal elbow.} For the integrated squared first derivative the minimax-optimal $\sqrt n$ regime begins at $s>2+d/4$ \citep{bickel1988estimating,laurent1996efficient}; it is attained only by higher-order (U-statistic) debiasing \citep{robins2017higher}. The plug-in condition \eqref{eq:smooth} is stronger, and is the honest condition for our estimator.

\subsection{Proof of Proposition~\ref{prop:degen}}
Let $j$ be null, so $g_j\equiv0$. In the expansion of Theorem~\ref{thm:clt}, (a)$=(\widehat P_X-P_X)0=0$ and (b)$=2\langle\widehat g_j-g_j,0\rangle=0$: the linear part is identically zero. Under $H_{0j}$, $f_0$ does not depend on $x_j$, and $A_j$ annihilates the part of $f_0(\Xb)$ constant in $x_j$ up to smoother bias, which is negligible by (A3). Hence $n\nuhat_j^{\mathrm{pl}}=\|A_j\bm\varepsilon\|^2(1+o_P(1))=\bm\varepsilon^\top A_j^\top A_j\bm\varepsilon(1+o_P(1))=n\,\bm\varepsilon^\top Q_j\bm\varepsilon(1+o_P(1))$, and the posterior-variance correction is $O_P(\epsilon_n'^2)$, lower order. With $Q_j=\sum_k\lambda_ku_ku_k^\top$, $\lambda_k\ge0$, and Gaussian $\bm\varepsilon$, $\bm\varepsilon^\top Q_j\bm\varepsilon=\sigma^2\sum_k\lambda_k\xi_k^2$ with $\xi_k=u_k^\top\bm\varepsilon/\sigma\stackrel{iid}{\sim}\mathcal N(0,1)$: a weighted $\chi^2_1$ mixture with mean $\sigma^2\mathrm{tr}Q_j>0$, non-Gaussian, and with a standardized law depending on $\{\lambda_k\}$ (non-pivotal). \hfill$\square$

\subsection{Proof of Lemma~\ref{lem:boot}}
Condition on the design and fitted hyperparameters. \emph{Step 1 (mis-scaling of the raw form).} $\bm r=(I-\mathsf S)\yv=(I-\mathsf S)f_0(\Xb)+(I-\mathsf S)\bm\varepsilon$; the first term is the smoothing bias, $o_P(1)$ coordinate-wise under (A3), and $\E\|(I-\mathsf S)\bm\varepsilon\|^2=\sigma^2\mathrm{tr}\{(I-\mathsf S)(I-\mathsf S)^\top\}=n\sigma^2\gamma_n$. Hence $\E^*[\bm e^\top\widetilde Q_j\bm e]=\mathrm{tr}\widetilde Q_j=\tfrac1n\sum_ir_i^2(A_j^\top A_j)_{ii}=\gamma_n\sigma^2\mathrm{tr}Q_j(1+o_P(1))$ by the law of large numbers over $i$ (the $r_i^2$ are asymptotically uncorrelated with the fixed diagonal weights), which is the stated mis-scaling.

\emph{Step 2 (two quadratic forms with a common normalized spectrum).} Write $U_n=\bm\varepsilon^\top Q_j\bm\varepsilon$ and $U_n^*=\bm e^\top\widetilde Q_j\bm e$, and their standardizations $Z_n=(U_n-\E U_n)/\mathrm{sd}(U_n)$, $Z_n^*=(U_n^*-\E^*U_n^*)/\mathrm{sd}^*(U_n^*)$. For a Gaussian vector and a symmetric PSD matrix $M$ with eigenvalues $\mu_k$, $\bm z^\top M\bm z\stackrel{d}{=}\sum_k\mu_k\xi_k^2$ with $\xi_k$ i.i.d.\ $\mathcal N(0,1)$, so $Z_n\stackrel{d}{=}\sum_k\bar\lambda_k(\xi_k^2-1)/\sqrt2$ with $\bar\lambda_k$ the normalized spectrum of $Q_j$ (the scale $\sigma^2$ cancels), and likewise $Z_n^*\stackrel{d}{=}\sum_k\bar\lambda_k^*(\xi_k^2-1)/\sqrt2$ with $\bar\lambda_k^*$ the normalized spectrum of $\widetilde Q_j$. Both standardized laws are therefore functions of the normalized spectra alone.

\emph{Step 3 (the normalized spectra agree).} $\widetilde Q_j=\tfrac1n R A_j^\top A_j R$ with $R=\mathrm{diag}(\bm r)$, and by Step~1 $R^2=\gamma_n\sigma^2 I+\Delta_n$ with $\Delta_n$ a diagonal matrix of mean-zero fluctuations. Writing $\widetilde Q_j=\gamma_n\sigma^2Q_j+E_n$, the perturbation $E_n=\tfrac1n(R A_j^\top A_j R-\gamma_n\sigma^2A_j^\top A_j)$ satisfies $\|E_n\|_F/\|\gamma_n\sigma^2Q_j\|_F\xrightarrow{P}0$ under (A3)--(A4) (the fluctuations $\Delta_n$ average out against the smooth kernel-derivative weights), so by Weyl/Hoffman--Wielandt the normalized spectra satisfy $\sum_k(\bar\lambda_k^*-\bar\lambda_k)^2\xrightarrow{P}0$.

\emph{Step 4 (invariance principle).} Under (S1), $\max_k\bar\lambda_k\to0$ and hence $\max_k\bar\lambda_k^*\to0$; the Lindeberg condition for the array $\{\bar\lambda_k(\xi_k^2-1)/\sqrt2\}$ holds and both $Z_n$ and $Z_n^*$ converge to $\mathcal N(0,1)$ \citep{dejong1987central}, giving the uniform convergence of distribution functions by P\'olya's theorem. Under (S2), $\ell^2$-convergence of the normalized spectra to $(\bar\lambda^\infty_k)$ implies, for the series $\sum_k\bar\lambda_k(\xi_k^2-1)$, convergence in $L^2$ and hence in distribution to $\sum_k\bar\lambda^\infty_k(\xi_k^2-1)$ (a fixed weighted-$\chi^2$ law with a continuous distribution function), for both $Z_n$ and $Z_n^*$ by Step~3; uniform convergence again follows from P\'olya's theorem. In either case the standardized bootstrap law and the standardized null law share a limit, which is the claim.

\emph{Step 5 (level).} By Proposition~\ref{prop:degen} the sampling null law of the studentized $T_j$ is that of $Z_n$ up to $o_P(1)$, and $T_j^{(b)}$ has the law of $Z_n^*$; since the bootstrap tail probability $p_j$ is computed from a consistent estimate of the null law of $T_j$, $\PP(p_j\le\alpha)\to\alpha$. \hfill$\square$

\begin{remark}[Which $d$, and when (A2$'$) is needed]\label{rem:scope}
In \eqref{eq:smooth} $d$ is the ambient input dimension $D$, because the contraction rate in (A3) is that of a $D$-variate function; when $f_0$ depends on only $s_0$ coordinates, an ARD kernel adapts and the effective dimension in the rate is $s_0$ \citep{yoo2016supremum,jiang2021variable}, which is why the $D=40$ experiment with $s_0=5$ active inputs is not in tension with \eqref{eq:smooth}. The boundary condition (A2$'$) enters only through the representer $h_j$ in Theorem~\ref{thm:clt}; at a null coordinate $\partial_jf_0\equiv0$ so both the interior representer and the boundary bracket of \S\ref{app:riesz} vanish, and the null calibration of Lemma~\ref{lem:boot} and Theorem~\ref{thm:boot} does not require (A2$'$).
\end{remark}

\begin{remark}[Data-dependent $Q_j$ and sample splitting]\label{rem:split}
$Q_j$ depends on the responses through the fitted hyperparameters, so conditioning on the design does not make it fixed. The proof above treats the hyperparameters as converging to a limit at a rate faster than the quadratic form's own fluctuations, which is standard for marginal-likelihood estimates \citep{kaufman2013role}. A theoretically cleaner variant fits the hyperparameters on an independent half of the sample and computes $\nuhat_j$ and the bootstrap on the other half, making $Q_j$ conditionally fixed. It keeps FDR control (Friedman $0.017$ vs.\ $0.022$ full-sample; misranking $0.022$ vs.\ $0.067$, $30$ replications) but pays the usual price of splitting when signals are weak: misranking power falls from $0.87$ to $0.57$ with only $n/2$ observations for inference (Table~\ref{tab:ablation}). We therefore use the full-sample procedure and offer splitting as an option in the code.
\end{remark}

\subsection{Proof of Theorem~\ref{thm:boot}}
(i) is Lemma~\ref{lem:boot}. (ii) Let $S^c$ be the null set, $|S^c|=m_0$, fixed. The vector $\bm T=(T_j)_{j\in S^c}$ consists of standardized quadratic forms $Z_{n,j}$ in the same Gaussian vector $\bm\varepsilon$, i.e.\ $\bm T\stackrel{d}{=}\big(\bm\varepsilon^\top\bar Q_j\bm\varepsilon-\mathrm{tr}\bar Q_j\big)_{j}$ with $\bar Q_j=Q_j/\|Q_j\|_F$; the bootstrap vector $\bm T^{(b)}$ has the same form with $\widetilde Q_j$ in place of $\sigma^2Q_j$. Joint convergence follows from the Cram\'er--Wold device: for any $\bm a\in\R^{m_0}$, $\bm a^\top\bm T$ is the standardized quadratic form with matrix $\sum_ja_j\bar Q_j$, to which Lemma~\ref{lem:boot} (with the spectral condition applied to that matrix) applies, and by Step~3 of its proof the bootstrap analogue shares its normalized spectrum. Hence $\bm T$ and $\bm T^{(b)}$ converge jointly to the same limit law $\bm L$. The map $\bm t\mapsto\max_jt_j$ is continuous, so $\max_jT_j\xrightarrow{d}\max_jL_j$ and $\max_jT_j^{(b)}\xrightarrow{d}\max_jL_j$ conditionally, and the bootstrap $(1-\alpha)$ quantile of the max converges to that of $\max_jL_j$ (whose distribution is continuous); rejecting when $\max_jT_j$ exceeds it therefore has asymptotic level $\alpha$ on the global null, and on a partial null the same argument restricted to $S^c$ gives FWER $\le\alpha$ by monotonicity of the max. (iii) By (i) each null $p_j$ is asymptotically super-uniform; the BY step-up controls FDR at $q\,m_0/D$ for arbitrarily dependent p-values with super-uniform nulls \citep{benjamini2001control}, giving $\mathrm{FDR}\le q\,m_0/D\le q$ in the limit. \hfill$\square$

\subsection{Proof of Proposition~\ref{prop:scale}}
With $\widetilde K$ the unit-variance kernel matrix, $K=\eta\widetilde K$ and $\fbar(\Xb)=\eta\widetilde K(\eta\widetilde K+\sigma^2I)^{-1}\yv=\widetilde K(\widetilde K+(\sigma^2/\eta)I)^{-1}\yv$, which depends on $(\eta,\sigma^2)$ only through $\sigma^2/\eta$; the same holds for $\widehat g_j(\xv)=(\partial_j\widetilde{\bm k}_{\xv})^\top(\widetilde K+(\sigma^2/\eta)I)^{-1}\yv$. Hence $\|\widehat g_j\|^2_{L^2(\widehat P_X)}$ is invariant under $(\eta,\sigma^2)\mapsto(c\eta,c\sigma^2)$. The posterior derivative variance is $\eta[\partial_j\partial_{j'}\widetilde k-(\partial_j\widetilde{\bm k})^\top(\widetilde K+(\sigma^2/\eta)I)^{-1}\partial_j\widetilde{\bm k}]$, linear in $\eta$ at fixed $\sigma^2/\eta$. \hfill$\square$

\subsection{Proof of Corollary~\ref{cor:select}}
Let $w_n$ be the level-$q$ rejection threshold on the $\nuhat$ scale; by Theorem~\ref{thm:boot}(iii) and Lemma~\ref{lem:boot}, $w_n\xrightarrow{P}0$. For $j\in S$, Proposition~\ref{prop:consist} gives $\nuhat_j\xrightarrow{P}\nu_j>0$, so if $\min_{j\in S}\nu_j/w_n\to\infty$ then $\PP(\nuhat_j>w_n\ \forall j\in S)\to1$ and every active coordinate is selected. For null $j$, (i) gives per-null rejection probability $\to0$ at the BY threshold and (iii) bounds the expected false fraction by $q$, so $\PP(\exists\text{ null }j\in\widehat S)\to0$. Combining, $\PP(\widehat S=S)\to1$. \hfill$\square$

\section{Algorithm and diagnostics}\label{app:alg}
\begin{algorithm}[h]
\caption{DPS: derivative-process sensitivity selection}\label{alg:dps}
\begin{algorithmic}[1]
\REQUIRE data $(\Xb,\yv)$; level $q$; bootstrap size $B$
\STATE fit SE-ARD GP; form $K_n^{-1}$ (one Cholesky)
\STATE $\nuhat_j\gets$ closed-form sensitivity \eqref{eq:nuhat} for all $j$
\STATE LOO residuals $r_i\gets\alpha_i/[K_n^{-1}]_{ii}$; maps $A_j\gets(\partial_j K_\star)K_n^{-1}$
\FOR{$b=1,\dots,B$}
\STATE draw $\bm e^{(b)}\sim\mathcal N(0,I_n)$; $\nuhat_j^{(b)}$ via \eqref{eq:boot}
\ENDFOR
\STATE studentize $\Rightarrow T_j$, p-values $p_j$
\STATE \textbf{return} $\widehat S=$ BY step-up on $\{p_j\}$ at level $q$ (or BH for more power)
\end{algorithmic}
\end{algorithm}

\begin{table}[h]
\centering\small
\caption{Diagnostics validating the theory (\S\ref{sec:theory}). Root-$n$ scaling for active scores (Thm.~\ref{thm:clt}); non-Gaussian positive-bias null (Prop.~\ref{prop:degen}); near-uniform null p-values (Thm.~\ref{thm:boot}).}
\label{tab:diag}
\adjustbox{max width=\linewidth}{%
\begin{tabular}{ll}
\toprule
Quantity & Measurement\\
\midrule
$\sqrt n\,\mathrm{sd}(\nuhat_j)$, active, $n{=}300/600/1200$ & $1.00\,/\,1.03\,/\,1.06$\\
bootstrap/MC sd ratio, active & $0.04$ (noise part only)\\
null bias decay exponent & $n^{-0.58}$ (not $o(n^{-1/2})$)\\
null Shapiro--Wilk $p$ & ${\sim}10^{-7}$ (non-Gaussian)\\
null p-value mean / KS stat & $0.46\,/\,0.08$\\
null $\PP(p\le0.05)$ / $\PP(p\le0.10)$ & $0.09\,/\,0.16$\\
\bottomrule
\end{tabular}}
\end{table}

\begin{figure}[h]
\centering
\includegraphics[width=\linewidth]{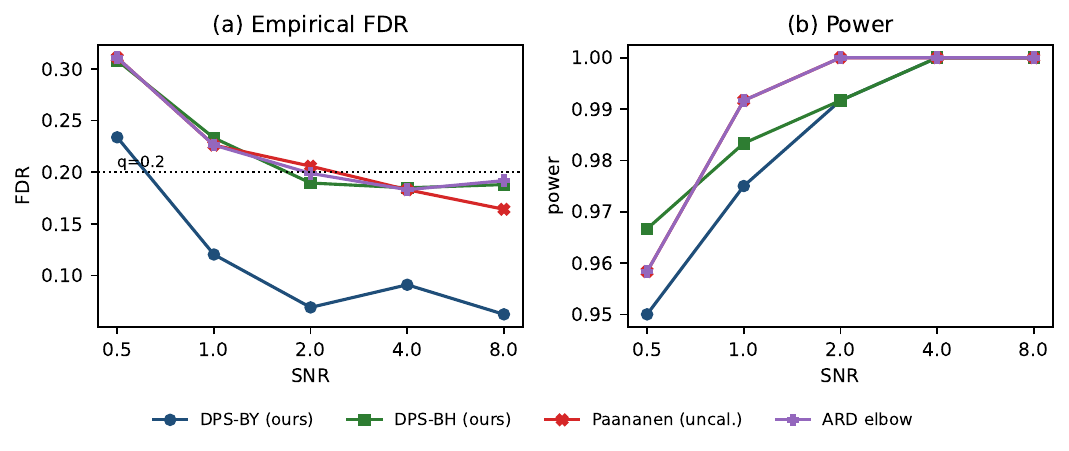}
\caption{FDR (left) and power (right) versus SNR on Friedman. DPS-BY's FDR falls below $q$ once SNR$\,\ge 1$ and stays controlled; the uncalibrated ranking remains above $q$ throughout. At extreme noise (SNR $0.5$) the asymptotic calibration degrades, an honest limitation.}
\label{fig:power}
\end{figure}

\textbf{Effect of signal strength.} Sweeping SNR on Friedman (Figure~\ref{fig:power}), DPS-BY controls FDR for SNR$\,\ge1$ ($0.12,0.07,0.09,0.06$ at SNR $1,2,4,8$) while the uncalibrated ranking sits at $0.16$--$0.23$, both at near-full power; at SNR $0.5$ the fit is too noisy for the asymptotic calibration and DPS-BY drifts to $0.23$, which we report rather than hide.

\section{Real-data details}\label{app:real}
We use three datasets and, in each, append planted exactly-null inputs so that a false-selection rate is measurable (Table~\ref{tab:real}; subsampled to $n\le500$ for exact GP fits, $8$--$14$ planted draws). On the \emph{red-wine quality} data of \citet{cortez2009modeling} ($n=1599$, $11$ physicochemical inputs) the derivative sensitivity ranks sulphates, alcohol, chlorides, total SO$_2$ and volatile acidity highest---the drivers the oenological literature identifies---with planted nulls at the floor ($\nuhat\approx0.01$); DPS-BY selects $0.1$ planted nulls per draw versus $1.5$ for both Paananen--KL and the ARD elbow, a $12\times$ reduction in spurious selections. The \emph{Appliances energy} data of \citet{candanedo2017data} ($n=19{,}735$, $27$ inputs of which two are authentic nulls) is a validity stress test: its authors deliberately included two random-noise inputs (\texttt{rv1}, \texttt{rv2}) as \emph{authentic} nulls, and the target is hard for a GP (leave-one-out $R^2\approx0.4$ at $n=450$--$900$). Here the marginal-likelihood fit overfits (lengthscales $0.2$--$0.6$, $\sigma^2\to10^{-8}$), in-sample residuals collapse, and the in-sample bootstrap becomes \emph{anti}-conservative (selecting nearly every input, nulls included). Replacing in-sample by leave-one-out residuals---closed-form for a GP, $r_i^{\mathrm{LOO}}=\alpha_i/[K_n^{-1}]_{ii}$ \citep{rasmussen2006gpml}, and never shrunk by the fit---restores validity: over six planted draws at $n=450$ DPS-BY never selects an authentic or planted null and selects only the strongest one or two inputs ($1.7$ per draw; RH$_1$, T$_6$), whereas Paananen--KL selects $3.3$ planted nulls per draw and the authentic nulls $17\%$ of the time and the ARD elbow $2.3$; in the well-fit regime the two residual choices agree (Friedman: FDR $0.017$ vs.\ $0.008$, power $1.0$ both; Table~\ref{tab:ablation}). We therefore recommend LOO residuals as the default. The classical diabetes data \citep{efron2004diabetes} behaves like wine (Appendix~\ref{app:extra}).

\textbf{Diagnostics behind the theory.} Appendix~\ref{app:alg} reports the measurements validating \S\ref{sec:theory}: root-$n$ scaling of active scores (Theorem~\ref{thm:clt}); the non-Gaussian, positively-biased null (Proposition~\ref{prop:degen}); and the two facts behind Lemma~\ref{lem:boot}---the raw bootstrap is mis-scaled by residual shrinkage (Kolmogorov distance $\approx0.3$) while the studentized bootstrap matches the null law ($\approx0.02$). Null p-values are near-uniform with a finite-sample tail inflation that shrinks with $n$ ($\PP(p_j\le0.05)=0.075$ at $n=300$, $0.062$ at $n=600$); BY absorbs the residue, which is why DPS-BY controls FDR throughout while DPS-BH is occasionally anticonservative. A stress test over $n$, $D$, SNR, and AR(1) correlation found DPS-BY controlling FDR in every cell (Appendix~\ref{app:extra}).

\begin{figure}[h]
\centering
\includegraphics[width=0.9\linewidth]{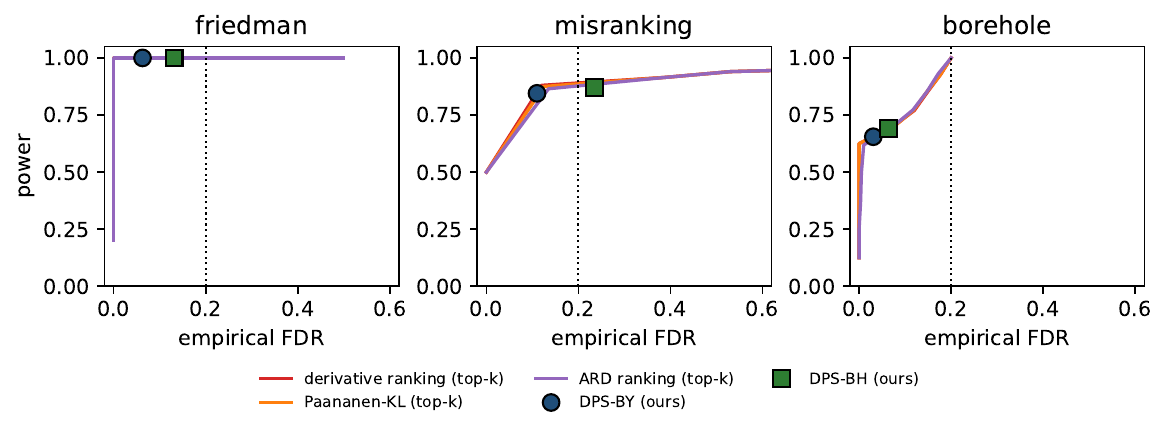}
\caption{\textbf{Power at matched FDR.} Sweeping the number of top-ranked coordinates traces an FDR--power curve for each uncalibrated ranking ($100$ replications). The calibrated DPS-BY point lies \emph{on} the curves in every benchmark: its lower raw power is FDR budget the rankings spend but are not entitled to, and it needs no oracle choice of $k$.}
\label{fig:matched}
\end{figure}

\begin{table}[h]
\centering\small
\caption{Ablations (DPS-BY, $q=0.2$). Higher dimension: Friedman with $35$ exact nulls. Bootstrap size: Friedman $D=10$. Residuals: LOO vs.\ in-sample.}
\label{tab:ablation}
\adjustbox{max width=\linewidth}{%
\begin{tabular}{l l cc}
\toprule
Ablation & Setting & FDR & Power\\
\midrule
Higher dimension & $D=40$, $n=400$, $16$ reps & $0.052\ (\pm0.020)$ & $0.99$\\
Bootstrap size & $B=100\,/\,300\,/\,1000$ ($20$ reps) & $0.008\,/\,0.008\,/\,0.008$ & $1.00$\\
Residuals (Friedman) & in-sample / LOO ($20$ reps) & $0.017\,/\,0.008$ & $1.00\,/\,1.00$\\
Residuals (misranking) & in-sample / LOO ($20$ reps) & $0.125\,/\,0.058$ & $0.88\,/\,0.88$\\
Sample split (Friedman) & full / split ($30$ reps) & $0.022\,/\,0.017$ & $1.00\,/\,1.00$\\
Sample split (misranking) & full / split ($30$ reps) & $0.067\,/\,0.022$ & $0.87\,/\,0.57$\\
\bottomrule
\end{tabular}}
\end{table}

\section{Additional experimental detail}\label{app:extra}
\begin{figure}[h]
\centering
\includegraphics[width=\linewidth]{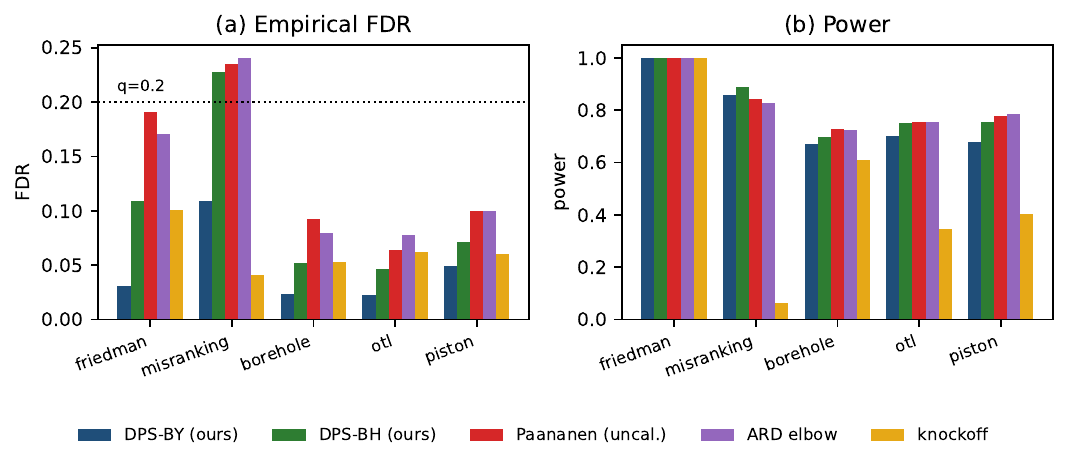}
\caption{Overview across five benchmarks at $q=0.2$ ($24$--$32$ replications; $100$-replication results with standard errors are in Table~\ref{tab:main} and, for the secondary error-controlled baselines, Table~\ref{tab:secondary}). ``Paananen (uncal.)'' here denotes the uncalibrated derivative ranking with a gap cutoff (DerivGap); the faithful Paananen--KL score behaves the same (Table~\ref{tab:main}). DPS-BY controls FDR on every benchmark with truly null inputs while the uncalibrated rankings breach repeatedly, at essentially the same power.}
\label{fig:benchmarks}
\end{figure}

\begin{table}[h]
\centering\small
\caption{Mat\'ern-5/2 kernel (closed-form derivative verified against finite differences to $3\times10^{-10}$; LOO residuals; 40/40/30 replications). Compare SE-ARD in Table~\ref{tab:main}.}
\label{tab:matern}
\adjustbox{max width=\linewidth}{%
\begin{tabular}{l cc cc}
\toprule
& \multicolumn{2}{c}{DPS-BY} & \multicolumn{2}{c}{DPS-BH}\\
\cmidrule(lr){2-3}\cmidrule(lr){4-5}
Benchmark & FDR (SE) & Power & FDR (SE) & Power\\
\midrule
Friedman & 0.013 (0.007) & 1.00 & 0.064 (0.015) & 1.00\\
misranking & 0.067 (0.024) & 0.88 & 0.187 (0.038) & 0.90\\
borehole & 0.010 (0.007) & 0.65 & 0.038 (0.017) & 0.69\\
\bottomrule
\end{tabular}}
\end{table}

\begin{table}[h]
\centering\small
\caption{A null input correlated with an active one ($n=300$, $D=10$, $4$ active; $30$ replications per $\rho$; latent correlation $\rho$, empirical correlation after the marginal transform in parentheses). ``Null'': fraction of runs selecting the correlated null. The target $\nu_j$ of that input is exactly zero.}
\label{tab:corr}
\adjustbox{max width=\linewidth}{%
\begin{tabular}{l cc cc cc}
\toprule
& \multicolumn{2}{c}{DPS-BY} & \multicolumn{2}{c}{Paananen--KL} & \multicolumn{2}{c}{ARD elbow}\\
\cmidrule(lr){2-3}\cmidrule(lr){4-5}\cmidrule(lr){6-7}
$\rho$ & FDR & Null & FDR & Null & FDR & Null\\
\midrule
0.5 (0.42) & 0.044 & 0.07 & 0.265 & 0.27 & 0.254 & 0.27\\
0.8 (0.75) & 0.051 & 0.10 & 0.267 & 0.27 & 0.256 & 0.27\\
0.95 (0.93) & 0.051 & 0.10 & 0.265 & 0.27 & 0.261 & 0.30\\
0.99 (0.99) & 0.053 & 0.10 & 0.266 & 0.27 & 0.245 & 0.30\\
\bottomrule
\end{tabular}}
\end{table}

\textbf{Active-coordinate CLT (Theorem~\ref{thm:clt}).} On a smooth above-elbow target ($f_0$ linear in the active coordinate, long-lengthscale fit), the centered estimator is root-$n$: $\sqrt n\cdot\mathrm{sd}(\nuhat_j)$ stabilizes at $1.00,\,1.03,\,1.06$ for $n=300,600,1200$ over $400$ Monte-Carlo datasets, and the studentized statistic passes a Shapiro--Wilk normality check. Consistent with the variance decomposition \eqref{eq:clt}, the residual multiplier bootstrap reproduces only the noise-driven component and therefore \emph{underestimates} the active-coordinate sampling SD (ratio $\approx0.04$ for a strong linear effect)---which is why the bootstrap is used for null calibration, not active-effect confidence intervals.

\textbf{Null degeneracy (Proposition~\ref{prop:degen}).} With a fixed well-specified kernel, the null-coordinate estimator is strictly positive with mean decaying as $n^{-0.58}$ (log-log slope over $n\in\{200,\dots,1600\}$, $250$ reps)---not $o(n^{-1/2})$---and is strongly non-Gaussian (Shapiro--Wilk $p\sim10^{-7}$), matching the weighted-$\chi^2$ prediction.

\textbf{Null p-value calibration (Theorem~\ref{thm:boot}(i)).} Pooling bootstrap p-values over truly null coordinates across $60$ datasets ($n=300$, $D=8$), the null p-values are approximately uniform (mean $0.46$; Kolmogorov--Smirnov statistic $0.08$) with a mild tail inflation, $\PP(p_j\le0.05)\approx0.09$ and $\PP(p_j\le0.10)\approx0.16$. The dependence-robust BY step-up absorbs this inflation, which is why DPS-BY controls FDR throughout while the BH variant is occasionally anticonservative.

\textbf{FDR control across regimes.} Beyond Table~\ref{tab:main}, the multiplier-bootstrap step-up was checked over $n\in\{200,300\}$, $D\in\{10,20\}$, SNR$\,\in\{5,10\}$, and AR(1) input correlation $\rho\in\{0,0.5\}$. DPS-BY controlled FDR at or below $q=0.2$ in every cell (range $0.075$--$0.181$); DPS-BH was occasionally mildly anticonservative (up to $0.28$), the expected consequence of cross-coordinate dependence from the shared fit.

\textbf{Settings.} GP hyperparameters were fit by L-BFGS marginal-likelihood maximization with two restarts; $B\in\{700,800,1000\}$ bootstrap draws; sensitivity evaluated at the training inputs (empirical measure). Replication counts ranged from $24$ to $32$ per benchmark. All derivative-process formulas are verified against finite differences in Appendix~\ref{app:verify}.
\section{Secondary baselines at 100 replications}\label{app:secondary}
Table~\ref{tab:secondary} repeats the error-controlled baselines at $100$ replications on the seeds of Table~\ref{tab:main}. All control FDR, but only DPS-BY is never less powerful: the inert-reference rule is erratic (power 0.06 on Friedman and 0.12 on misranking, but 0.64 on borehole), and knockoffs and HSIC match DPS-BY's power on Friedman at roughly twice its FDR yet collapse on the weak misranking effect.

\begin{table}[h]
\centering\small
\caption{Error-controlled baselines at $100$ replications, same seeds and data as Table~\ref{tab:main} (FDR / power, $q=0.2$). GP fits use the same marginal likelihood with analytic gradients (identical optimum and DPS-BY selections to the finite-difference fitter in $12/12$ checks).}
\label{tab:secondary}
\adjustbox{max width=\linewidth}{%
\begin{tabular}{l cccc}
\toprule
Benchmark & DPS-BY & ARD inert-dummy & Knockoffs & HSIC\\
\midrule
Friedman & 0.063 / 1.00 & 0.000 / 0.06 & 0.120 / 0.99 & 0.116 / 0.98\\
misranking & 0.110 / 0.85 & 0.006 / 0.12 & 0.065 / 0.10 & 0.130 / 0.54\\
borehole & 0.031 / 0.66 & 0.028 / 0.64 & 0.051 / 0.62 & 0.049 / 0.52\\
\bottomrule
\end{tabular}}
\end{table}

\section{Family-wise error of the max statistic}\label{app:fwer}
Theorem~\ref{thm:boot}(ii) is asymptotic; Table~\ref{tab:fwer} measures it. With in-sample residuals the single-step max-$T$ procedure is mildly liberal at these sample sizes (FWER $0.130$ at $\alpha=0.10$), the family-wise analogue of the null tail inflation in Table~\ref{tab:diag}; leave-one-out residuals, our recommended default, bring it to nominal ($0.080$ at $D=10$ and $0.083$ at $D=40$) with power $1.00$ throughout.

\begin{table}[h]
\centering\small
\caption{Family-wise error of the max-statistic procedure (Theorem~\ref{thm:boot}(ii)), $\alpha=0.10$, Friedman with $D-5$ exact nulls.}
\label{tab:fwer}
\adjustbox{max width=\linewidth}{%
\begin{tabular}{l c cc}
\toprule
Setting (residuals) & Reps & FWER (SE) & Power\\
\midrule
$D=10$, $n=300$ (in-sample) & 100 & 0.130 (0.034) & 1.00\\
$D=10$, $n=600$ (in-sample) & 38 & 0.132 (0.055) & 1.00\\
$D=10$, $n=300$ (LOO) & 100 & 0.080 (0.027) & 1.00\\
$D=40$, $n=400$ (LOO) & 24 & 0.083 (0.056) & 1.00\\
\bottomrule
\end{tabular}}
\end{table}

\section{Large $n$: block-averaged DPS}\label{app:largen}
\textbf{A naive inducing-point approximation breaks calibration.} Replacing the exact posterior by the inducing-point (subset-of-regressors) mean, which is also the predictive mean of the variational approximation of \citet{titsias2009variational}, reproduces $\nuhat_j$ almost exactly (correlation $0.9995$ with the exact estimate, $n=800$, $m=300$) yet its bootstrap test selected $2$--$4$ null inputs in every one of $5$ replications, against $0$--$1$ for exact DPS. The low-rank fit represents the signal with larger, less regular weights, and that approximation error leaks into the derivatives along null inputs; it is a bias the residual bootstrap, which models only noise, cannot see.

\textbf{Block averaging preserves the structure of Lemma~\ref{lem:boot}.} Fit the hyperparameters on a held-out subset; split the remaining data into $K$ disjoint blocks; compute $\nuhat_{jk}$ and its bootstrap draws by exact DPS within each block (independent multipliers per block, LOO residuals); average both over blocks and studentize. Conditional on the held-out hyperparameters, under $H_{0j}$ the averaged statistic is $\bm\varepsilon^\top Q_j\bm\varepsilon$ with the block-diagonal $Q_j=K^{-1}\mathrm{blockdiag}(Q_{j1},\dots,Q_{jK})$, so Lemma~\ref{lem:boot} and Theorem~\ref{thm:boot} apply verbatim. Moreover the spectrum of $Q_j$ is the union of the block spectra scaled by $1/K$, so for blocks with comparable spectra the largest normalized eigenvalue is $O(K^{-1/2})$: condition (S1) holds automatically as the number of blocks grows, and the null limit is Gaussian. The cost is $O(n\,n_b^2)$ for blocks of size $n_b$, linear in $n$.

\textbf{Evidence.} On Friedman with $D=20$ ($15$ exact nulls) block-averaged DPS gave FDR $0.028$ at power $1.00$ at $n=4{,}000$ ($6$ replications) and selected exactly the five active inputs at every $n$ from $2{,}000$ to $20{,}000$ (Table~\ref{tab:largen}), taking $15$\,s at $n=20{,}000$ on one core. On the full Appliances data ($n=19{,}735$, two authentic and six planted nulls; two draws of the planted nulls and block split, about $39$\,s each) it selected no null input in either draw, and $11$ and $3$ real inputs respectively, with RH$_1$, T$_6$ and RH$_8$ in both; power varies with the held-out hyperparameter fit on this hard target. The ARD elbow at the same hyperparameters admitted $4$ and $3$ planted nulls, and in the second draw both authentic ones.

\begin{table}[h]
\centering\small
\caption{Block-averaged DPS on Friedman with $D=20$ ($15$ exact nulls), blocks of $500$, hyperparameters fit on a held-out $500$ points; single CPU core. Time is dominated by the hyperparameter fit.}
\label{tab:largen}
\adjustbox{max width=\linewidth}{%
\begin{tabular}{r c cc cc}
\toprule
$n$ & Blocks & Total (s) & Hyper.\ (s) & FDP & Power\\
\midrule
2{,}000 & 3 & 4.6 & 4.1 & 0.00 & 1.00\\
5{,}000 & 9 & 6.7 & 5.1 & 0.00 & 1.00\\
10{,}000 & 19 & 14.8 & 11.4 & 0.00 & 1.00\\
20{,}000 & 39 & 15.3 & 8.2 & 0.00 & 1.00\\
\bottomrule
\end{tabular}}
\end{table}

\section{Selection for Bayesian optimization}\label{app:bo}
Table~\ref{tab:bo} asks whether selecting inputs before Bayesian optimization helps. With $n_0=100$ initial points DPS-BY certified no input in any run (10/10 empty selections) and fell back to full-$D$ BO; the gap rules selected inputs aggressively, often dropped truly active ones, and lost badly (mean final regret $0.351$ for the ARD elbow against $0.123$). With $n_0=200$ DPS-BY engaged and made no false inclusions over the ten runs (ARD elbow: $34$), but it usually missed one weak active input, and full-$D$ BO with an ARD kernel was best overall. The lesson is one of scope: FDR-calibrated selection maximizes precision---the right objective when the selected set is itself the scientific claim---whereas dimension reduction for optimization rewards recall. For BO we recommend a recall-oriented rule or no reduction; DPS is useful there chiefly as a guard, declining to reduce when the evidence is weak.

\begin{table}[h]
\centering\small
\caption{Selection before Bayesian optimization: Hartmann-6 hidden in $D=30$ ($6$ random active inputs per run), selection on the initial design, then $30$ EI iterations on the selected inputs; $10$ runs per setting. ``Sel.'': mean inputs selected; ``TP'': mean true actives among them; ``FP'': total false inclusions over the runs. An empty selection falls back to full-$D$ BO (same rule for every method).}
\label{tab:bo}
\adjustbox{max width=\linewidth}{%
\begin{tabular}{l ccc c}
\toprule
Method & Sel. & TP & FP & Final regret (SE)\\
\midrule
\multicolumn{5}{l}{\emph{$n_0=100$ initial points}}\\
Oracle & 6.0 & 6.0 & 0 & 0.090 (0.019)\\
DPS-BY & 0.0 & 0.0 & 0 & 0.123 (0.022)\\
ARD elbow & 7.8 & 4.6 & 32 & 0.351 (0.124)\\
Paananen--KL & 7.9 & 4.7 & 32 & 0.345 (0.125)\\
Full-D & -- & -- & -- & 0.123 (0.022)\\
Random & -- & -- & -- & 1.064 (0.113)\\
\midrule
\multicolumn{5}{l}{\emph{$n_0=200$ initial points}}\\
Oracle & 6.0 & 6.0 & 0 & 0.103 (0.023)\\
DPS-BY & 4.5 & 4.5 & 0 & 0.185 (0.042)\\
DPS-BH & 5.1 & 5.0 & 1 & 0.203 (0.049)\\
ARD elbow & 8.8 & 5.4 & 34 & 0.151 (0.041)\\
Paananen--KL & 10.5 & 5.4 & 51 & 0.165 (0.038)\\
Full-D & -- & -- & -- & 0.090 (0.020)\\
Random & -- & -- & -- & 1.020 (0.110)\\
\bottomrule
\end{tabular}}
\end{table}

\end{document}